\documentclass{bmvc2k}

\title{A Revisit to the Decoder for Camouflaged Object Detection}

\addauthor{Seung Woo Ko$^*$}{seungwoo.ko@lgresearch.ai}{1,4}
\addauthor{Joopyo Hong$^*$}{dals2539@snu.ac.kr}{1}
\addauthor{Suyoung Kim$^*$}{ksyo96@snu.ac.kr}{2}
\addauthor{Seungjai Bang$^*$}{epdl6403@snu.ac.kr}{3,5}
\addauthor{Sungzoon Cho}{zoon@snu.ac.kr}{3}
\addauthor{Nojun Kwak}{nojunk@snu.ac.kr}{2}
\addauthor{Hyung-Sin Kim$^\dagger$}{hyungkim@snu.ac.kr}{1}
\addauthor{Joonseok Lee$^\dagger$}{joonseok@snu.ac.kr}{1,6}

\addinstitution{
 Graduate School of Data Science,\\
 Seoul National University\\
 Seoul, Korea
}
\addinstitution{
 Department of Intelligence and Information, 
 Seoul National University\\
 Seoul, Korea
}
\addinstitution{
 Department of Industrial Engineering, \\
 Seoul National University\\
 Seoul, Korea
}
\addinstitution{
 LG AI Research\\
 Seoul, Korea
}
\addinstitution{
 LG Electronics\\
 Seoul, Korea
}
\addinstitution{
 Google Research \\
 Mountain View, CA, USA\\
 \hspace{-0.25cm} $^*$Equal contribution \\
 \hspace{-0.25cm} $^\dagger$Corresponding authors
}

\runninghead{Ko et al.}{A Revisit to the Decoder for Camouflaged Object Detection}

\usepackage{indentfirst}

\usepackage{comment,enumitem}
\usepackage{color}
\usepackage{wrapfig}
\usepackage{mathtools}
\usepackage{tabularx}
\usepackage{multirow}
\usepackage{tabularray}
\usepackage{supertabular}
\usepackage{caption}
\usepackage{graphicx}
\usepackage{booktabs}

\usepackage{xcolor}

\newcommand\seungwoo[1]{\textcolor{orange}{#1}}

\newcommand\red[1]{\textcolor{red}{#1}}
\newcommand\nj[1]{\textcolor{purple}{#1}}

\newcommand\proposal{ENTO}
\newcommand\predec{Enrich Decoder}
\newcommand\maindec{base decoder}
\newcommand\postdec{Retouch Decoder}

\AtEndPreamble{
    \usepackage[capitalize]{cleveref}
    \crefname{section}{Sec.}{Secs.}
    \Crefname{section}{Section}{Sections}
    \crefname{table}{Tab.}{Tabs.}
    \Crefname{table}{Table}{Tables}
}

\begin{document}

\maketitle


\begin{abstract}
Camouflaged object detection (COD) aims to generate a fine-grained segmentation map of camouflaged objects hidden in their background.
Due to the hidden nature of camouflaged objects, it is essential for the decoder to be tailored to effectively extract proper features of camouflaged objects and extra-carefully generate their complex boundaries.
In this paper, we propose a novel architecture that augments the prevalent decoding strategy in COD with \predec\ and \postdec, which help to generate a fine-grained segmentation map.
Specifically, the \predec\ amplifies the channels of features that are important for COD using channel-wise attention.
\postdec\ further refines the segmentation maps by spatially attending to important pixels, such as the boundary regions.
With extensive experiments, we demonstrate that \proposal\ shows superior performance using various encoders, with the two novel components playing their unique roles that are mutually complementary.
\end{abstract}
\section{Introduction}
\label{sec:intro}

\begin{figure}[t]
  \centering
  \includegraphics[width=0.49\linewidth]{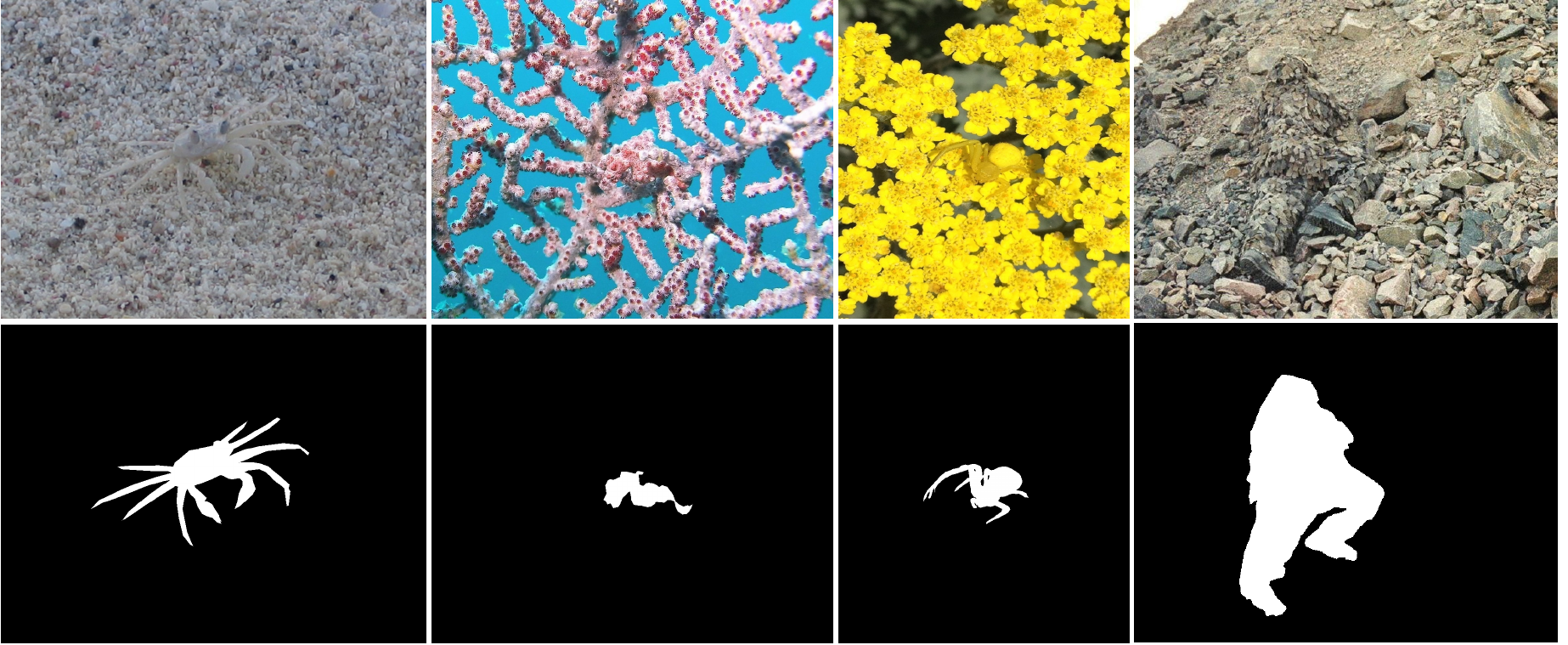}
  \includegraphics[width=0.49\linewidth]{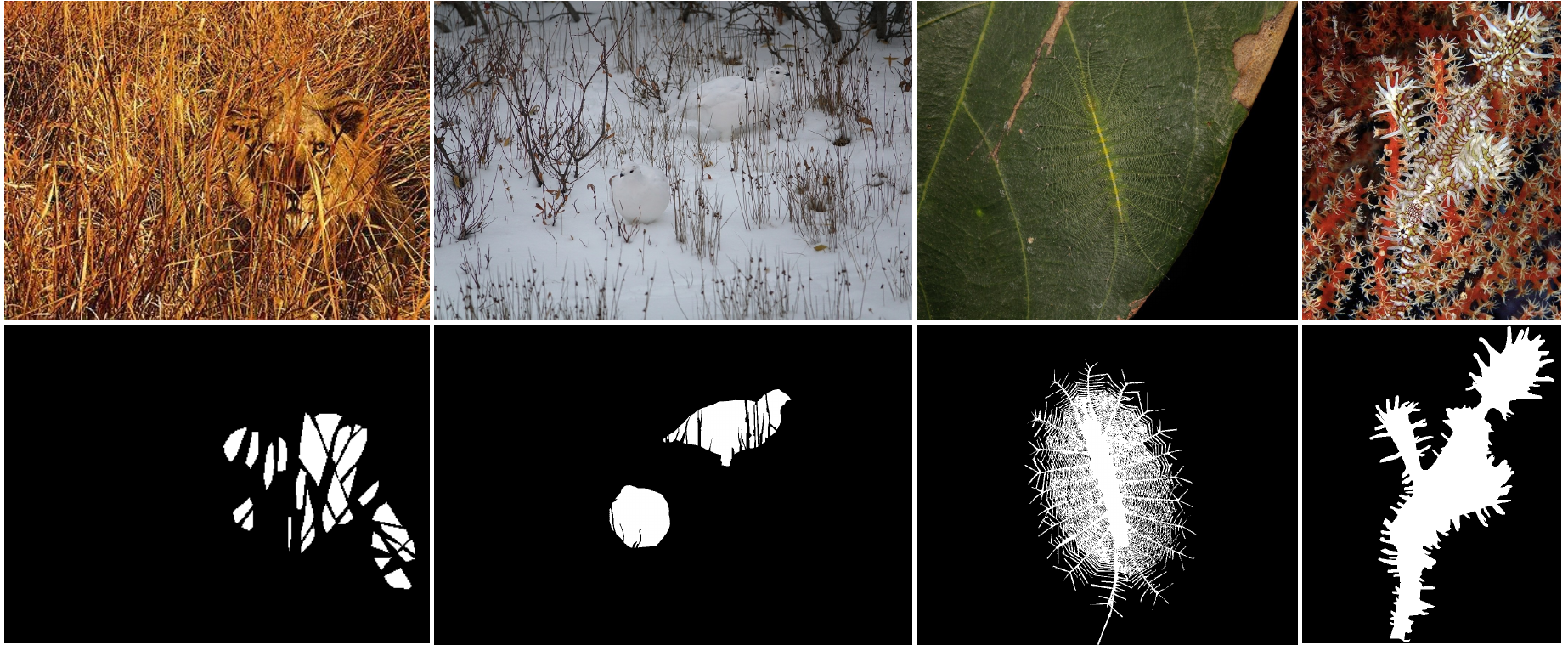}
  \vspace{0.2cm}
  \caption{\textbf{Examples of camouflaged objects in COD10K and NC4K.} 
  }
  \label{fig:cod_example}
\end{figure}

Object segmentation is a widely researched topic in computer vision.
Of its several branches, Camouflaged Object Detection (COD)
\cite{fan2020sinet} targets images that contain naturally or artificially camouflaged objects and aims to correctly \textit{segment}\footnote{Precisely speaking, this is an image segmentation task, where the expected output is a pixel-wise segmentation map, not bounding boxes. We use this idiomatic expression interchangeably.} them from the background.
COD is applied to polyp segmentation~\cite{fan2020pranet} in medicine, surface defect detection in manufacturing~\cite{bhajantri2006camouflage}, enemy detection in military~\cite{liu2023mhnet}, and camouflaged organisms in ecology~\cite{merilaita2017howcamouflage}.

Camouflaged objects bring a variety of challenges in segmentation. \cref{fig:cod_example} illustrates several representative examples: using protective coloring to disturb others' visual recognition, hiding behind another object (occlusion), or exhibiting complex shapes.
To segment these objects accurately, the model needs not only to precisely understand the given scene and individual objects, but also to be equipped with an exceptional segmentation map generator,
surpassing the level of general object segmentation models.

\begin{table*}[t]
\centering
\resizebox{\textwidth}{!}{%
\begin{tabular}{lllllllllllll}
\toprule
\multicolumn{1}{l|}{\multirow{2}{*}{\textbf{Methods}}} & \multicolumn{4}{c|}{\textbf{COD10K (2,026)}} & \multicolumn{4}{c|}{\textbf{NC4K (4,121)}} & \multicolumn{4}{c}{\textbf{CAMO (250)}} \\ 
\multicolumn{1}{c|}{} & ${S_\alpha}$ ↑ & ${F_\beta^w}$ ↑ & ${E_\phi}$ ↑ & \multicolumn{1}{c|}{${M}$ ↓} & ${S_\alpha}$ ↑ & ${F_\beta^w}$ ↑ & ${E_\phi}$ ↑ & \multicolumn{1}{c|}{${M}$ ↓} & ${S_\alpha}$ ↑ & ${F_\beta^w}$ ↑ & ${E_\phi}$ ↑ & ${M}$ ↓ \\ \midrule

\multicolumn{1}{l|}{ZoomNet~\cite{pang2022zoomnet}} & \textbf{0.870} & 0.782 & 0.912 & \multicolumn{1}{l|}{0.023} & \textbf{0.884} & 0.829 & \textbf{0.922} & \multicolumn{1}{l|}{\textbf{0.034}} & \textbf{0.865} & \textbf{0.812} & \textbf{0.914} & \textbf{0.052} \\ 

\multicolumn{1}{l|}{HitNet~\cite{hu2023hitnet}} & 0.867 & \textbf{0.803} & \textbf{0.926} & \multicolumn{1}{l|}{\textbf{0.022}} & 0.872 & \textbf{0.832} & 0.921 & \multicolumn{1}{l|}{0.037} & 0.836 & 0.798 & 0.893 & 0.060 \\

\bottomrule
\end{tabular}}
\vspace{0.2cm}
\caption{\textbf{Comparison of baseline models using Transformer encoder.} 
ZoomNet with the same encoder setting as HitNet shows comparable or even better results for some metrics.}
\label{tab:intro_comparison}
\end{table*}

Numerous methods have been proposed for this task, primarily focusing on the extraction of rich features to distinguish inconspicuous objects from their surroundings \cite{hu2023hitnet,huang2023fspnet,liu2022dtinet}.
It has been demonstrated that using high-resolution input images is effective for COD~\cite{hu2023hitnet, xing2023sarnet}, more than that for general object segmentation~\cite{zeng2019towards,parra2022inputeffect,ha2024nemo}.
However, has the performance improvements using such powerful encoding tools been matched with an equally effective decoding strategy?
To verify this, we replace the CNN encoder backbone in ZoomNet \cite{pang2022zoomnet} with a Transformer backbone, HitNet \cite{hu2023hitnet}.
The results in \cref{tab:intro_comparison} indicates that the older ZoomNet easily reaches or even surpasses the performance achieved by HitNet by simply taking its Transformer-based encoder and higher resolution input.
This implies that there has been little advances in the decoding architecture, and there is room for improvement in the decoder structure to fully take advantage of the encoding capacity advanced by recent works.

In this regards, we propose to equip the base decoder \cite{fan2021sinetv2} with two additional processes, before and after it.
Before the base decoder, we \emph{enrich the image features} so they are more suitable for camouflaged objects, since the regular image features extracted by the encoder may not be sufficiently informative. 
Based on the enriched features and a coarse prediction map from this pre-decoding step, the base decoder generates a refined segmentation map. 
After the decoding, the segmentation map is \emph{retouched again} since it may still not be perfect, especially regarding the complex shapes of camouflaged objects. 

Reflecting this idea, we propose \textbf{\proposal}, equipped with the \textbf{EN}rich and Re\textbf{TO}uch Decoders that sandwich the base decoder, specially designed for camouflaged object segmentation.
The core in \proposal\ is a step-wise generation of the segmentation map through three consecutive decoding steps.
Specifically, the first step (\predec) is to reorganize the image features by channel-wise attention and combination of multi-level features, focusing on the features (\emph{e.g.}, texture and shape) critical for the COD task.
The enriched features are then fed into the base decoder~\cite{fan2021sinetv2} to output segmentation maps at different scales. 
The last \postdec\ further enhances the boundaries of the objects by applying spatial attention to focus on those regions of the segmentation map.
These pre- and post-steps enable more precise representation of object boundaries while keeping the overall structure.


%

\begin{figure}
    \centering
    \includegraphics[width=0.48\linewidth]{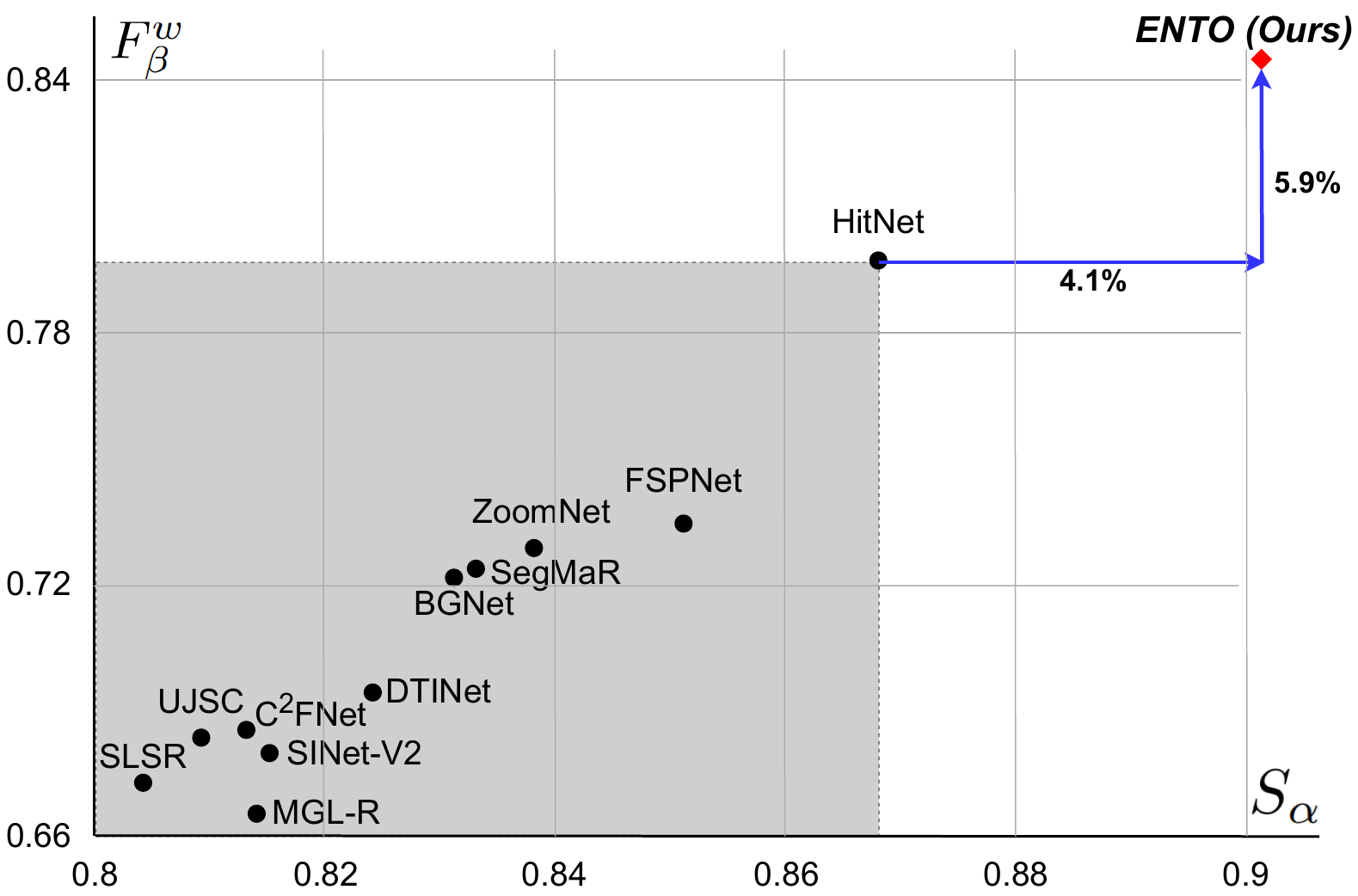}
    \vspace{0.2cm}
    \caption{\textbf{Performances in two representative metrics, $S_\alpha$ and $F^w_\beta$, on COD10K~\cite{fan2020sinet}.}}
    \label{fig:performance}
\end{figure}

Our comprehensive experiments verify that \proposal\ achieves state-of-the-art performance across multiple representative COD datasets, as in \cref{fig:performance}. 
We also show through extensive experiments that the proposed decoder structure is able to adapt to any type of feature encoder at any image resolution, surpassing previous state-of-the-art models.
\section{Related Work}
\label{sec:related}

\vspace{0.1cm} \noindent
\textbf{Camouflaged Object Detection (COD).}
 The COD task was first proposed as a sub-task of object segmentation by Fan \emph{et al.}~\cite{fan2020sinet}, providing a large-scale dataset (COD10K). 
Since then, numerous models have been proposed to tackle this problem~\cite{kim2019deep,lyu2021rank, chen2022bgnet, zhu2022bsanet, hu2023hitnet, huang2023fspnet, fan2021sinetv2}.
Since camouflaged objects are harder to segment, auxiliary tasks are often applied, such as object ranking~\cite{lyu2021rank} and edge detection~\cite{chen2022bgnet}. 
Some works are inspired by biology.
SINet~\cite{fan2020sinet}, for example, takes an idea from predators that first search for prey in a general sense and subsequently identify its precise location.
BSA-Net~\cite{zhu2022bsanet} mimics how human discovers camouflaged objects using a two-stream attention network.

\vspace{0.1cm} \noindent
\textbf{Input and Encoding.} 
Recently, high resolution images have been utilized for COD, taking advantage of richer information on the boundary regions.
HitNet~\cite{hu2023hitnet} utilizes high resolution images to better capture high-frequency details.
SARNet~\cite{xing2023sarnet} improves the segmentation quality by training on higher resolution images.
To extract fine-grained features from intricate images, Transformers~\cite{vaswani2017attention} have been adopted in COD.
Pyramid Vision Transformer (PVT)~\cite{wang2022pvt} improves the vanilla ViT~\cite{dosovitskiy2020vit} by providing a pyramid-style encoding structure, extracting features in a similar manner to CNNs.
Recent COD models adopt ViT~\cite{huang2023fspnet} and PVT~\cite{hu2023hitnet,xing2023sarnet} as the feature extractor.

\vspace{0.1cm} \noindent
\textbf{Decoding Strategy.}
The U-Net architecture \cite{ronneberger2015unet} has been extensively utilized for intricate object segmentation domains, such as medical imaging and remote sensing.
Drawing inspiration from SINet \cite{fan2020sinet}, a majority of studies on COD are rooted in this U-shaped decoding, sequentially aggregating high to low-level features with skip connections.
Some recent works have tried to enhance decoding strategies for COD.
PreyNet~\cite{zhang2022preynet} densely aggregates features from neighboring layers during the high-to-low decoding phase.
HitNet~\cite{hu2023hitnet} employs a feedback refinement between layers.
FSPNet~\cite{huang2023fspnet} employs a pyramidal shrinkage decoding strategy that progressively aggregates adjacent features. Furthermore, several studies~\cite{mei2021pfnet,zhu2022bsanet} employ a coarse-to-fine decoding strategy, initially generating a coarse map followed by refinement stages that aim to capture information grounded in the coarse map.

Despite studies on decoding strategies, they remain relatively unexplored compared to the recent significant advances in encoders.
In particular, while the performance of decoders based on CNN-based encoders is almost saturated, they fall short in effectively leveraging the rich feature maps produced by advanced Transformer-based encoders like ViT~\cite{huang2023fspnet} and PVT~\cite{hu2023hitnet,xing2023sarnet}.
In response to this, we introduce an innovative decoding strategy that augments the prevalent decoding structure with preprocessing and postprocessing decoders that help to enhance the features necessary for COD and improve the boundary details.

\section{Preliminary}
\label{sec:prelim}

\noindent
\textbf{Problem Formulation and Notations.}
The Camouflaged Object Detection (COD) is formulated in the same way as a regular image segmentation problem, except that the target object is camouflaged and not easily seen at a glance.
The input is an image $\mathbf{X} \in \mathbb{R}^{H \times W \times 3}$ with RGB channels, where $H$ and $W$ stand for the height and width of the image.
For each image, the COD model needs to generate a bitmap $\mathbf{\hat{Y}} \in \{0, 1\}^{H \times W}$, predicting the ground truth map $\mathbf{Y} \in \{0, 1\}^{H \times W}$, where 1 indicates the pixel belongs to the camouflaged object and 0 otherwise.
For simplicity, we do not distinguish the identity of an object, even if there are multiple camouflaged objects in the same scene.

\vspace{0.1cm} \noindent
\textbf{Our Base Architecture.}
We take a common encoder-decoder structure for image segmentation, where the encoder extracts essential features to segment the target object, and the decoder generates a segmentation map corresponding to it.
We adopt a particular encoder and decoder architecture described below, which is widely used for modern COD models.


\vspace{0.1cm} \noindent
\textbf{Feature Encoder.}
Given an input image $\mathbf{X}$, the encoder $E$ extracts features $\mathbf{f}_i$ with different resolutions at each level $i$, where $i = 1, ..., L$.
$E$ may accommodate various types of backbones, \emph{e.g.}, CNNs, or Transformers.
As most image encoders generate different numbers of channels at each level, we make the channel size the same across the layers by additionally applying a convolution layer. 
We denote the resulting feature map by $\mathbf{f}'_i$, where $i = 1, ..., L$.

\begin{figure*}
  \centering
  \includegraphics[width=1.0\linewidth]{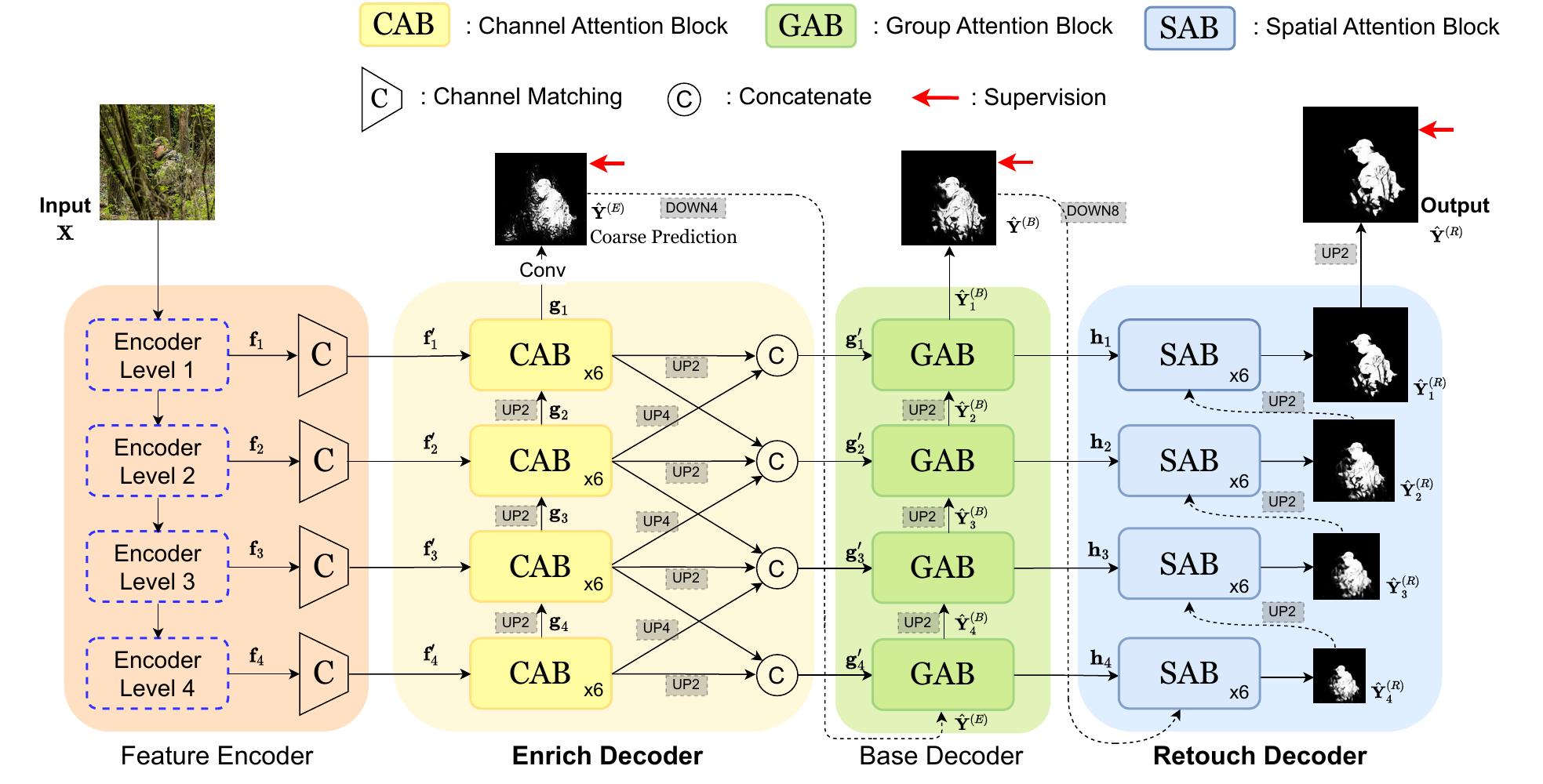}
  \vspace{0.1cm}
  \caption{\textbf{Overall Architecture of \proposal, comprising a feature encoder and the three consecutive decoders.} We show the architecture with 4 levels ($L=4$), consistent with our full model, but the number of levels $L$ can be adjusted according to the feature encoder.}
  \label{fig:architecture}
\end{figure*}

\vspace{0.1cm} \noindent
\textbf{Base Decoder.}
In this work, we propose pre- and post-decoding steps on an existing decoding structure, which we call base decoder.
Aligned with the encoder, we adopt a common multi-level decoder to fully take advantage of the multi-level feature maps.
As illustrated in \cref{fig:architecture}, we take $L$ levels of Group Attention Blocks (GABs) inspired by \cite{fan2021sinetv2}.
GAB is a residual learning process employing the group guidance operation, focusing more on important information about objects through attention with guidance from the prior segmentation map and gradually improving the map through a sequence of $S$
group attentions.
It expands the segmentation map step by step from low to high resolution, learning rough and abstract patterns at the low resolution while finer details at the higher resolution. 
The operations of GAB are detailed in \cref{sec:gab}.
Although such a multi-level decoder structure is prevalent in COD, our \maindec\ slightly deviates from \cite{fan2021sinetv2} in three ways: 1) we remove the reverse operation on the guidance maps, 2) we expand each GAB to take four group operations, and 3) we provide an  additional GAB stage to fully utilize the features from the encoder.

\section{The Proposed Method: \proposal}
\label{sec:method}

On top of the base encoder-decoder structure in \cref{sec:prelim}, we introduce two novel steps that make the segmentation model more suitable for camouflaged objects, as illustrated in \cref{fig:architecture}.
Specifically, we insert a pre-decoding step (\predec) between the encoder and the base decoder to adapt the encoded image features more suitable for COD, and a post-decoding step (\postdec) at the end, to refine the produced segmentation map, focusing on the object boundaries.

The overall decoding structure is as follows.
Taking the encoded image features at $L$ different scales, \predec selectively amplifies the channels that contain important cues for detecting camouflaged objects.
Taking an attention strategy used for image super-resolution, this step benefits to prepare a particular set of features (\emph{e.g.}, texture and shape) that are more important for camouflaged objects.
Additionally, features at different resolutions are fused so that the \maindec\ can utilize both coarse and fine information simultaneously at each level.
Then, the \maindec\ adds details step-by-step, refining the low-resolution segmentation map using higher resolution features produced by the previous step.
We use Group Attention Blocks (GABs) \cite{fan2021sinetv2}, suitable to handle our multi-resolution features, but any image decoder can be used for this step.
Lastly, the \postdec\ further enhances the object boundaries of the produced segmentation map.
We adopt spatial attention to amplify the signals of the edges and detailed areas that may not have been fully segmented by the base decoder.

\subsection{Pre-Decoding Step: \predec}
\label{sec:method:predecoder}

In order to effectively utilize the encoded features $\mathbf{f}'_i$, we propose a novel pre-decoding step to adapt the features towards the camouflaged objects, before the base decoder.
Specifically, we apply a module that has been shown effective in image super-resolution; namely, channel attention block (CAB) \cite{zhang2018cab}.
This module selectively emphasizes particular feature channels important for super-resolution by channel-wise attention to the image features.
Such a mechanism would benefit the COD task similarly, since some different sets of features (\emph{e.g.}, texture and shape) may be more important than others (\emph{e.g.}, color) to detect camouflaged objects, unlike regular ones.

\begin{figure}
  \centering
  \includegraphics[width=0.44\linewidth, trim=2.3cm 0cm 0cm 0cm]{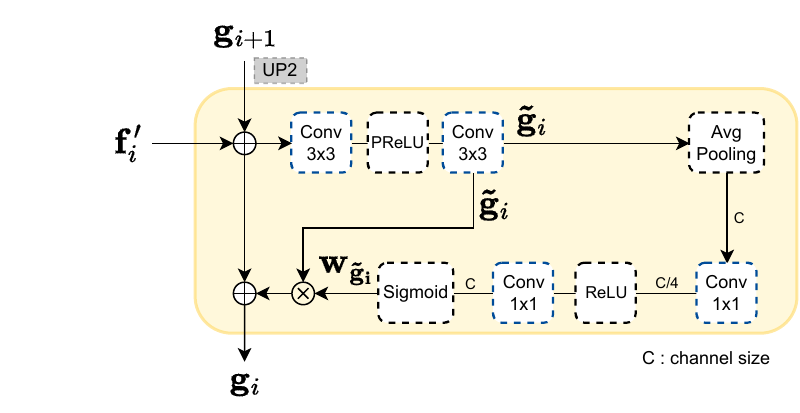}
  \includegraphics[width=0.54\linewidth, trim=2.3cm 0cm 0cm 0cm]{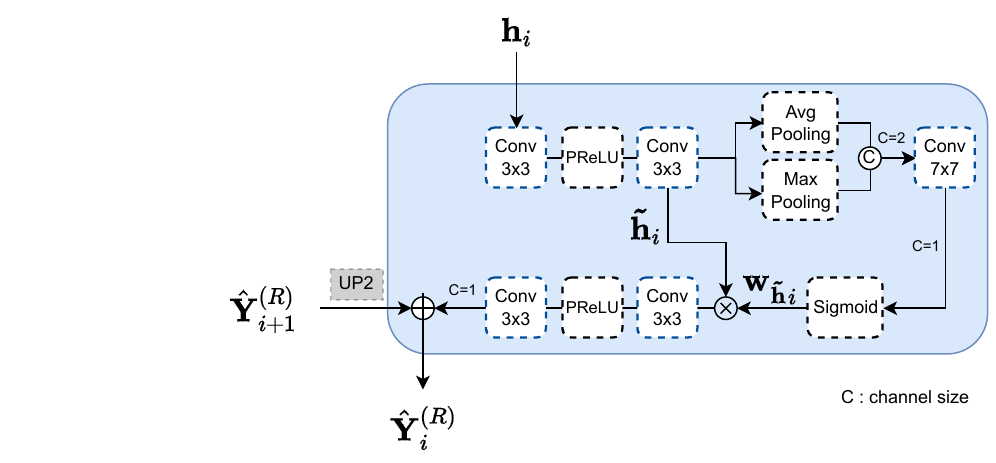}
  \vspace{0.1cm}
  \caption{\textbf{Channel Attention Block (CAB)~\cite{zhang2018cab} and Spatial Attention Block (SAB).}}
  \label{fig:CAB_SAB}
\end{figure}

As illustrated in \cref{fig:CAB_SAB}(left), the channel attention block (CAB) takes two inputs: the general feature map $\mathbf{f}'_i$ at the $i$-th level, and the adapted feature map $\mathbf{g}_{i+1}$ by the previous CAB.
At each level $i$, CAB upsamples $\mathbf{g}_{i+1}$ (to match the dimensionality with $\mathbf{f}_i'$), adds them with $\mathbf{f}_i'$, takes channel-wise attention to reogranize the information, and residually adds the input features back to the output $\mathbf{g}_i$.  
Formally, CAB at level $i$ performs:
\begin{align}
  \Tilde{\mathbf{g}}_{i} &= \mathtt{Conv3}\circ\mathtt{PReLU}\circ\mathtt{Conv3}(\mathbf{f}'_{i} \oplus \mathbf{g}_{i+1}),
  \quad
  \mathbf{w}_{\Tilde{\mathbf{g}_i}} = \sigma[\mathtt{Conv1}\circ\mathtt{ReLU}\circ\mathtt{Conv1}(\mathtt{Pool}(\Tilde{\mathbf{g}}_{i}))] \nonumber \\
  \mathbf{g}_{i} &= (\mathbf{w}_{\Tilde{\mathbf{g}_i}} \otimes \Tilde{\mathbf{g}}_{i}) \oplus (\mathbf{f}'_{i} \oplus \mathbf{g}_{i+1})
\label{eq:CAB}
\end{align}
where $i = 1, ..., L$, and $\mathbf{w}_{\Tilde{g}_i} \in (0,1)^C$ is the channel attention weights.
$\mathtt{Conv3}$, $\mathtt{Conv1}$, and $\mathtt{Pool}$ stand for $3 \times 3$ and $1 \times 1$ convolutions, and global average pooling, respectively.
$\oplus$ and $\otimes$ indicate element-wise addition and channel-wise multiplication, respectively.
At the highest level $L$, CAB simply performs channel-wise attention on $\mathbf{f}'_L$, taking only $\mathbf{f}'_L$ as input (that is, $\mathbf{g}_{L+1} = 0$).


Then, how does this CAB produce features more suitable for camouflage objects?
At the lowest layer, \predec\ produces a coarse segmentation map $\hat{\mathbf{Y}}^{(E)}$, as illustrated in \cref{fig:architecture}, with an additional convolution. 
By applying a loss on the difference between this coarse segmentation map and the ground truth, we guide the \predec\ to learn a set of feature maps suitable for the camouflage objects.
We also use this coarse map as the starting point of the base decoder, so the loss applied in the subsequent decoders indirectly affects \predec\ to learn COD-specific features as well.
In this way, the features important to COD are selectively amplified.




Lastly, the \predec\ concatenates the produced feature $\mathbf{g}_i$ with that on one level lower and higher ($\mathbf{g}_{i-1}$ and $\mathbf{g}_{i+1}$).
These aggregated features are appropriately upsampled, concatenated, and reverted back to the original channel size using convolution to produce a single feature map $\mathbf{g}'_i$.
This multi-level fusion allows the \maindec\ to access useful information scattered at different resolutions.

\subsection{Post-Decoding Step: \postdec}
\label{sec:method:postdecoder}

In addition to the pre-decoding step, we also add a post-decoding step to further improve the quality of the segmentation map produced by the base decoder.
Particularly, the \postdec\ performs fine-tuned enhancement, focusing on object boundaries, through spatial attention on segmentation maps starting from the final output from the \maindec. 
Similarly to other previous steps, \postdec\ progressively generates expanding prediction maps starting from the lowest-resolution.

Each \postdec\ at the $i$-th level applies multiple spatial attention blocks (SAB) that take the features $\mathbf{h}_i$ from the \maindec\ and the segmentation map from the previous SAB stage.
As shown in \cref{fig:CAB_SAB}(right), each SAB generates spatial weights $\mathbf{w}_{\Tilde{\mathbf{h}}_i}$ indicating which pixels should be further modified from the previous map.
Applying convolutions on the features weighted by this generates an improved map, added to the higher level map $\hat{\mathbf{Y}}^{(R)}_{i+1}$ to produce enhanced segmentation map $\hat{\mathbf{Y}}^{(R)}_{i}$.
Overall, SAB at the $i$-th level operates
 \begin{align}
  \Tilde{\mathbf{h}}_{i} &= \mathtt{Conv3}\circ\mathtt{PReLU}\circ\mathtt{Conv3}(\mathbf{h}_i),
  \quad
  \mathbf{w}_{\Tilde{\mathbf{h}}_i} = \sigma(\mathtt{Conv7}\circ \mathtt{Pool}(\Tilde{\mathbf{h}}_i)) \nonumber
  \\
  \hat{\mathbf{Y}}^{(R)}_{i} &= [\mathtt{Conv3}\circ\mathtt{PReLU}\circ\mathtt{Conv3}(\mathbf{w}_{\Tilde{\mathbf{h}}_i} \otimes \Tilde{\mathbf{h}}_{i})] \oplus \hat{\mathbf{Y}}^{(R)}_{i+1}
  \label{eq:SAB}
\end{align}
where $i = 1, ..., L$ and $\hat{\mathbf{Y}}^{(R)}_{L+1} = \hat{\mathbf{Y}}^{(B)}$.
$\Tilde{\mathbf{h}}_i$ is the intermediate feature with the same size as $\mathbf{h}_{i}$, and $\mathbf{w}_{\Tilde{h}_i} \in (0,1)^{H\times W}$ is the spatial attention weights.
$\mathtt{Pool}$ indicates a concatenation of average and max pooled features channel-wise; that is, $[\mathtt{AVG_c}(\Tilde{\mathbf{h}}_i), \mathtt{MAX_c}(\Tilde{\mathbf{h}}_i)]$.
Through these operations, SAB ultimately learns to identify the pixels that need modifications through spatial attention and provides residual learning to the output segmentation maps.

\subsection{Training Objectives}

\proposal\ sequentially refines a segmentation map throughout the decoding steps, supervising to the coarse map generated by the \predec\ and to each of the final output maps from base decoder and \postdec.
\cref{fig:architecture} illustrates where we apply supervision.
In each case, the output map is first upscaled to the input resolution, then we apply the following two common loss functions for object detection: pixel-level weighted binary cross entropy (wBCE)~\cite{qin2019basnet} and weighted intersection-over-union (wIOU) loss~\cite{wei2020f3net} to account for the overall overlap of the output map with the ground truth.
For both losses, each pixel is weighted to reflect its difficulty to be detected.
The loss at each decoding step is given by
\begin{equation}  \mathcal{L}(\hat{\mathbf{Y}},\mathbf{Y}) = \mathcal{L}_\text{wBCE}(\mathtt{UP}(\hat{\mathbf{Y}}),\mathbf{Y}) + \mathcal{L}_\text{wIOU}(\mathtt{UP}(\hat{\mathbf{Y}}),\mathbf{Y})\label{eq:loss}
\end{equation}
and by combining them, the overall loss is given by
\begin{equation}
  \mathcal{L} = \underbrace{\mathcal{L}(\hat{\mathbf{Y}}^{(E)}, \mathbf{Y})}_{\text{\predec}} + \underbrace{\mathcal{L}(\hat{\mathbf{Y}}^{(B)}, \mathbf{Y})}_{\text{\maindec}} + \underbrace{\mathcal{L}(\hat{\mathbf{Y}}^{(R)}, \mathbf{Y})}_{\text{\postdec}},
  \label{eq:losstotal}
\end{equation}
where $\hat{\mathbf{Y}}^{(E,B,R)}$ are output segmentation maps at each decoder, and $\mathbf{Y}$ is the ground truth segmentation map.
$\mathtt{UP}$ is upscaling function to the input image size using bilinear interpolation.


%
\section{Experiments}
\label{sec:experiments}

\subsection{Experiment Setup}
\label{sec:exp:setup}

\noindent
\textbf{Datasets.}
We evaluate on three widely-used COD datasets: COD10K \cite{fan2020sinet}, CAMO \cite{le2019camo} and NC4K \cite{lyu2021rank}. COD10K includes 5,066 camouflaged, 3,000 background, and 1,934 non-camouflaged images. CAMO consists of 1,250 camouflaged and 1,250 non-camouflaged images. NC4K consists of 4,121 images containing camouflaged objects from the Internet.
We train the model using only camouflaged images from COD10K and CAMO train sets (4,040 images) and evaluate on NC4K, COD10K, and CAMO test sets.
For evaluation metrics, we use

\vspace{0.1cm} \noindent
\textbf{Evaluation Metrics.} 
We evaluate with
four common metrics: S-measure (${S_\alpha}$) \cite{fan2017structure}, weighted F-measure (${F_\beta^w}$) \cite{margolin2014wfmeasure}, mean E-measure (${E_\phi}$) \cite{fan2021emeasure}, and mean absolute error (${M}$) \cite{perazzi2012mae}.
S-measure quantifies the structural similarity between the model output and the ground truth, which is important in COD tasks that usually contain complex shapes of objects.
Weighted F-measure is a modified version of F-measure that provides more reliable evaluation.
Mean E-measure quantifies the pixel-level matching and image-level statistics between the predicted output and the ground truth.
Mean absolute error directly quantifies the error in each pixel value averaged over the whole image.
See \cref{sec:impl_details} for more experimental settings.

\subsection{Comparison to Existing Methods}
\label{sec:exp:comparison}

\noindent
\textbf{Quantitative Comparison.}
As reported in \cref{tab:all_results}, our proposed method outperforms all baseline methods in all metrics. 
In particular, on COD10K, \proposal\ shows of 4.15\%, 5.89\%, and 1.72\% improvement in ${S_\alpha}$, ${F_\beta^w}$, and ${E_\phi}$ from the next best model, HitNet \cite{hu2023hitnet}.
Similarly, on NC4K, \proposal\ improves 2.84\% in ${S_\alpha}$ and 4.73\% in ${F_\beta^w}$ from the second-best models.
These results demonstrate that our model competently segments camouflaged objects.

\begin{table*}
\centering
\resizebox{\textwidth}{!}{%
\begin{tabular}{lcllllllllllll}
\toprule
\multicolumn{1}{l|}{\multirow{2}{*}{\textbf{Methods}}} & \multicolumn{1}{c|}{\multirow{2}{*}{Publications}} & \multicolumn{4}{c|}{\textbf{COD10K (2,026)}} & \multicolumn{4}{c|}{\textbf{NC4K (4,121)}} & \multicolumn{4}{c}{\textbf{CAMO (250)}} \\ 
\multicolumn{1}{l|}{} & \multicolumn{1}{c|}{} & ${S_\alpha}$ ↑ & ${F_\beta^w}$ ↑ & ${E_\phi}$ ↑ & \multicolumn{1}{c|}{${M}$ ↓} & ${S_\alpha}$ ↑ & ${F_\beta^w}$ ↑ & ${E_\phi}$ ↑ & \multicolumn{1}{c|}{${M}$ ↓} & ${S_\alpha}$ ↑ & ${F_\beta^w}$ ↑ & ${E_\phi}$ ↑ & ${M}$ ↓ \\ \midrule
\multicolumn{1}{l|}{SINet~\cite{fan2020sinet}} & \multicolumn{1}{c|}{$\text{CVPR}_{20}$} & 0.771 & 0.551 & 0.806 & \multicolumn{1}{l|}{0.051} & 0.808 & 0.723 & 0.871 & \multicolumn{1}{l|}{0.058} & 0.751 & 0.606 & 0.771 & 0.100 \\
\multicolumn{1}{l|}{SLSR~\cite{lv2021slsr}} & \multicolumn{1}{c|}{$\text{CVPR}_{21}$} & 0.804 & 0.673 & 0.880 & \multicolumn{1}{l|}{0.037} & 0.840 & 0.766 & 0.895 & \multicolumn{1}{l|}{0.048} & 0.787 & 0.696 & 0.838 & 0.080 \\
\multicolumn{1}{l|}{MGL-R~\cite{zhai2021mgl}} & \multicolumn{1}{c|}{$\text{CVPR}_{21}$} & 0.814 & 0.666 & 0.852 & \multicolumn{1}{l|}{0.035} & 0.833 & 0.740 & 0.867 & \multicolumn{1}{l|}{0.052} & 0.775 & 0.673 & 0.812 & 0.088 \\
\multicolumn{1}{l|}{PFNet~\cite{mei2021pfnet}} & \multicolumn{1}{c|}{$\text{CVPR}_{21}$} & 0.800 & 0.660 & 0.877 & \multicolumn{1}{l|}{0.040} & 0.829 & 0.745 & 0.888 & \multicolumn{1}{l|}{0.053} & 0.782 & 0.695 & 0.842 & 0.085 \\
\multicolumn{1}{l|}{UJSC~\cite{li2021ujsc}} & \multicolumn{1}{c|}{$\text{CVPR}_{21}$} & 0.809 & 0.684 & 0.884 & \multicolumn{1}{l|}{0.035} & 0.842 & 0.771 & 0.898 & \multicolumn{1}{l|}{0.047} & 0.800 & 0.728 & 0.859 & 0.073 \\
\multicolumn{1}{l|}{$\text{C}^{2}$FNet~\cite{sun2021c2fnet}} & \multicolumn{1}{c|}{$\text{IJCAI}_{21}$} & 0.813 & 0.686 & 0.890 & \multicolumn{1}{l|}{0.036} & 0.838 & 0.762 & 0.897 & \multicolumn{1}{l|}{0.049} & 0.796 & 0.719 & 0.854 & 0.080 \\
\multicolumn{1}{l|}{SINet-V2~\cite{fan2021sinetv2}} & \multicolumn{1}{c|}{$\text{TPAMI}_{22}$} & 0.815 & 0.680 & 0.887 & \multicolumn{1}{l|}{0.037} & 0.847 & 0.770 & 0.903 & \multicolumn{1}{l|}{0.048} & 0.820 & 0.743 & 0.882 & 0.070 \\
\multicolumn{1}{l|}{BGNet~\cite{chen2022bgnet}} & \multicolumn{1}{c|}{$\text{IJCAI}_{22}$} & 0.831 & 0.722 & 0.901 & \multicolumn{1}{l|}{0.033} & 0.851 & 0.788 & 0.907 & \multicolumn{1}{l|}{0.044} & 0.812 & 0.749 & 0.870 & 0.073 \\
\multicolumn{1}{l|}{DTINet~\cite{liu2022dtinet}} & \multicolumn{1}{c|}{$\text{ICPR}_{22}$} & 0.824 & 0.695 & 0.896 & \multicolumn{1}{l|}{0.034} & 0.863 & 0.792 & \textcolor{blue}{\textbf{0.917}} & \multicolumn{1}{l|}{\textcolor{blue}{\textbf{0.041}}} & \textcolor[rgb]{0,0.502,0}{\textbf{0.857}} & 0.796 & \textcolor[rgb]{0,0.502,0}{\textbf{0.916}} & \textcolor[rgb]{0,0.502,0}{\textbf{0.050}} \\
\multicolumn{1}{l|}{SegMaR~\cite{jia2022segmar}} & \multicolumn{1}{c|}{$\text{CVPR}_{22}$} & 0.833 & 0.724 & 0.899 & \multicolumn{1}{l|}{0.034} & 0.841 & 0.781 & 0.896 & \multicolumn{1}{l|}{0.046} & 0.815 & 0.753 & 0.874 & 0.071 \\
\multicolumn{1}{l|}{ZoomNet~\cite{pang2022zoomnet}} & \multicolumn{1}{c|}{$\text{CVPR}_{22}$} & 0.838 & 0.729 & \textcolor{blue}{\textbf{0.911}} & \multicolumn{1}{l|}{0.029} & 0.853 & 0.784 & 0.912 & \multicolumn{1}{l|}{0.043} & 0.820 & 0.752 & 0.883 & 0.066 \\ 
\multicolumn{1}{l|}{FEDER-R2N~\cite{he2023fedder}} & \multicolumn{1}{c|}{$\text{CVPR}_{23}$} & 0.844 & $\; \; \; \;$- & \textcolor{blue}{\textbf{0.911}} & \multicolumn{1}{l|}{0.029} & 0.862 & $\; \; \; \;$- & 0.913 & \multicolumn{1}{l|}{0.042} & 0.836 & $\; \; \; \;$- & 0.897 & 0.066 \\
\multicolumn{1}{l|}{FSPNet~\cite{huang2023fspnet}} & \multicolumn{1}{c|}{$\text{CVPR}_{23}$} & \textcolor{blue}{\textbf{0.851}} & \textcolor{blue}{\textbf{0.735}} & 0.895 & \multicolumn{1}{l|}{\textcolor{blue}{\textbf{0.026}}} & \textcolor[rgb]{0,0.502,0}{\textbf{0.879}} & \textcolor{blue}{\textbf{0.816}} & 0.915 & \multicolumn{1}{l|}{0.048} & \textcolor{blue}{\textbf{0.856}} & \textcolor{blue}{\textbf{0.799}} & 0.899 & \textcolor[rgb]{0,0.502,0}{\textbf{0.050}} \\

\multicolumn{1}{l|}{HitNet~\cite{hu2023hitnet}} & \multicolumn{1}{c|}{$\text{AAAI}_{23}$} & \textcolor[rgb]{0,0.502,0}{\textbf{0.868}} & \textcolor[rgb]{0,0.502,0}{\textbf{0.798}} & \textcolor[rgb]{0,0.502,0}{\textbf{0.932}} & \multicolumn{1}{l|}{\textcolor[rgb]{0,0.502,0}{\textbf{0.024}}} & \textcolor{blue}{\textbf{0.870}} & \textcolor[rgb]{0,0.502,0}{\textbf{0.825}} & \textcolor[rgb]{0,0.502,0}{\textbf{0.921}} & \multicolumn{1}{l|}{\textcolor[rgb]{0,0.502,0}{\textbf{0.039}}} & 0.844 & \textcolor[rgb]{0,0.502,0}{\textbf{0.801}} & \textcolor{blue}{\textbf{0.902}} & \textcolor{blue}{\textbf{0.057}} \\
\multicolumn{1}{l|}{\textbf{ENTO (Ours)}} & \multicolumn{1}{c|}{-} & \textcolor{red}{\textbf{0.904}} & \textcolor{red}{\textbf{0.845}} & \textcolor{red}{\textbf{0.948}} & \multicolumn{1}{l|}{\textcolor{red}{\textbf{0.018}}} & \textcolor{red}{\textbf{0.904}} & \textcolor{red}{\textbf{0.864}} & \textcolor{red}{\textbf{0.942}} & \multicolumn{1}{l|}{\textcolor{red}{\textbf{0.029}}} & \textcolor{red}{\textbf{0.881}} & \textcolor{red}{\textbf{0.841}} & \textcolor{red}{\textbf{0.928}} & \textcolor{red}{\textbf{0.047}} \\ \bottomrule
\end{tabular}}
\vspace{0.2cm}
\caption{\textbf{Overall comparison on COD datasets.} 
The \textcolor{red}{\textbf{1st}}, \textcolor[rgb]{0,0.502,0}{\textbf{2nd}}, \textcolor{blue}{\textbf{3rd}} best are highlighted.}
\label{tab:all_results}
\end{table*}

\begin{table*}
\centering
\resizebox{\textwidth}{!}{%
\begin{tabular}{lccccc|cc|cc|cc}
\toprule

\multicolumn{1}{c|}{\multirow{2}{*}{\textbf{Encoder Backbone}}} & \multicolumn{1}{c|}{\multirow{2}{*}{\textbf{Resolution}}} & \multicolumn{1}{c|}{\multirow{2}{*}{\textbf{Best Baseline}}} & \multicolumn{1}{c|}{\multirow{2}{*}{\textbf{Publications}}} & \multicolumn{2}{c|}{${S_\alpha} \uparrow$} & \multicolumn{2}{c|}{${F_\beta^w} \uparrow$} & \multicolumn{2}{c|}{${E_\phi} \uparrow$} & \multicolumn{2}{c}{${M} \downarrow$} \\ 
\multicolumn{1}{c|}{} & \multicolumn{1}{c|}{} & \multicolumn{1}{c|}{} & \multicolumn{1}{c|}{} & 
Ours & Base & Ours & Base & Ours & Base & Ours & Base \\ \midrule

\multicolumn{1}{c|}{\multirow{2}{*}{PVTv2-B2~\cite{wang2022pvt}}} & \multicolumn{1}{c|}{$352\times352$} & \multicolumn{1}{c|}{HitNet~\cite{hu2023hitnet}} & \multicolumn{1}{c|}{$\text{AAAI}_{23}$} & \textbf{0.889} & 0.827 & \textbf{0.835} & 0.727 & \textbf{0.932} & 0.907 & 0.032 & \textbf{0.029} \\
\multicolumn{1}{c|}{ } & \multicolumn{1}{c|}{$704\times704$} & \multicolumn{1}{c|}{HitNet~\cite{hu2023hitnet}} & \multicolumn{1}{c|}{$\text{AAAI}_{23}$} & \textbf{0.900} & 0.870 & \textbf{0.857} & 0.825 & \textbf{0.937} & 0.921 & \textbf{0.031} & 0.039 \\ \midrule

\multicolumn{1}{c|}{ViT~\cite{dosovitskiy2020vit}} & \multicolumn{1}{c|}{$384\times384$} & \multicolumn{1}{c|}{FSPNet~\cite{huang2023fspnet}} & \multicolumn{1}{c|}{$\text{CVPR}_{23}$} & \textbf{0.886} & 0.879 & \textbf{0.834} & 0.816 & \textbf{0.935} & 0.915 & \textbf{0.033} & 0.048 \\ \midrule

\multicolumn{1}{c|}{\multirow{2}{*}{Res2Net50~\cite{fan2020sinet}}} & \multicolumn{1}{c|}{$352\times352$} & \multicolumn{1}{c|}{SINet-V2~\cite{fan2021sinetv2}} & \multicolumn{1}{c|}{$\text{PAMI}_{22}$} & \textbf{0.854} & 0.847 & \textbf{0.775} & 0.770 & 0.901 & \textbf{0.903} & \textbf{0.046} & 0.048 \\
\multicolumn{1}{c|}{ } & \multicolumn{1}{c|}{$384\times384$} & \multicolumn{1}{c|}{FEDER-R2N~\cite{he2023fedder}} & \multicolumn{1}{c|}{$\text{CVPR}_{23}$} & \textbf{0.864} & 0.862 & 0.802 & -- & \textbf{0.913} & \textbf{0.913} & \textbf{0.041} & 0.042 \\ \midrule 

\multicolumn{1}{c|}{ResNet50~\cite{he2016resnet}} & \multicolumn{1}{c|}{$576\times576$} & \multicolumn{1}{c|}{ZoomNet~\cite{pang2022zoomnet}}& \multicolumn{1}{c|}{$\text{CVPR}_{22}$} & \textbf{0.858} & 0.853 & \textbf{0.786} & 0.784 & \textbf{0.914} & 0.912 & \textbf{0.043} & \textbf{0.043} \\ \bottomrule

\end{tabular}}
\vspace{0.2cm}
\caption{\textbf{Comparison using same encoder and resolution settings as baselines on NC4K dataset.} We select the best baseline models using various encoder backbone and resolutions, and report our model's performance under that setting. The better result is boldfaced.}
\label{tab:impartial_comparison}
\end{table*}

\begin{table}[t]
\centering
\resizebox{\textwidth}{!}{%
\begin{tabular}{lcllllllllllll}
\toprule
\multicolumn{1}{l|}{\multirow{2}{*}{\textbf{Methods}}} & \multicolumn{1}{c|}{\multirow{2}{*}{Publications}} & \multicolumn{4}{c|}{\textbf{COD10K (2,026)}} & \multicolumn{4}{c|}{\textbf{NC4K (4,121)}} & \multicolumn{4}{c}{\textbf{CAMO (250)}} \\ 
\multicolumn{1}{l|}{} & \multicolumn{1}{c|}{} & ${S_\alpha}$ ↑ & ${F_\beta^w}$ ↑ & ${E_\phi}$ ↑ & \multicolumn{1}{c|}{${M}$ ↓} & ${S_\alpha}$ ↑ & ${F_\beta^w}$ ↑ & ${E_\phi}$ ↑ & \multicolumn{1}{c|}{${M}$ ↓} & ${S_\alpha}$ ↑ & ${F_\beta^w}$ ↑ & ${E_\phi}$ ↑ & ${M}$ ↓ \\ \midrule

\multicolumn{1}{l|}{ZoomNet~\cite{pang2022zoomnet}} & \multicolumn{1}{c|}{$\text{CVPR}_{22}$} & 0.870 & 0.782 & 0.912 & \multicolumn{1}{l|}{0.023} & 0.884 & 0.829 & 0.922 & \multicolumn{1}{l|}{0.034} & 0.865 & 0.812 & 0.914 & 0.052 \\ 

\multicolumn{1}{l|}{HitNet~\cite{hu2023hitnet}} & \multicolumn{1}{c|}{$\text{AAAI}_{23}$} & 0.867 & 0.803 & 0.926 & \multicolumn{1}{l|}{0.022} & 0.872 & 0.832 & 0.921 & \multicolumn{1}{l|}{0.037} & 0.836 & 0.798 & 0.893 & 0.060 \\

\multicolumn{1}{l|}{\textbf{ENTO (Ours)}} & \multicolumn{1}{c|}{-} & \textbf{0.894} & \textbf{0.827} & \textbf{0.944} & \multicolumn{1}{l|}{\textbf{0.021}} & \textbf{0.895} & \textbf{0.849} & \textbf{0.935} & \multicolumn{1}{l|}{\textbf{0.033}} & \textbf{0.882} & \textbf{0.837} & \textbf{0.928} & \textbf{0.046} \\ \bottomrule
\end{tabular}}
\vspace{0.2cm}
\caption{\textbf{Comparison to baseline models using Transformer encoder.} 
}
\label{tab:trans_results}
\end{table}

\vspace{0.1cm} \noindent
\textbf{Impartial Comparison.}
As a high-performance encoder and high-resolution input images evidently lead to better performance~\cite{hu2023hitnet, xing2023sarnet, zeng2019towards, parra2022inputeffect}, a fair comparison should be conducted with the baselines; \emph{e.g.}, same encoder backbone and input image resolution.
\cref{tab:impartial_comparison} compares the performance of our model with the best-performing baseline models under various combinations of encoder backbones and resolutions. Our innovative decoding strategy achieves the best performance in most metrics with the same backbone and resolutions as baselines.

In \cref{tab:trans_results}, we compare our model with the same Transformer backbone (PVTv2-b2) and input resolution (768), similar to the results shown in \cref{tab:intro_comparison}. ENTO significantly outperforms both models using Transformer backbone showing that it matches the advances in encoder performance, such as using Transformer-base backbones, while the 
previous models have not fully utilized them.

\begin{figure*}[t]
  \centering
  \includegraphics[width=0.98\linewidth]
  {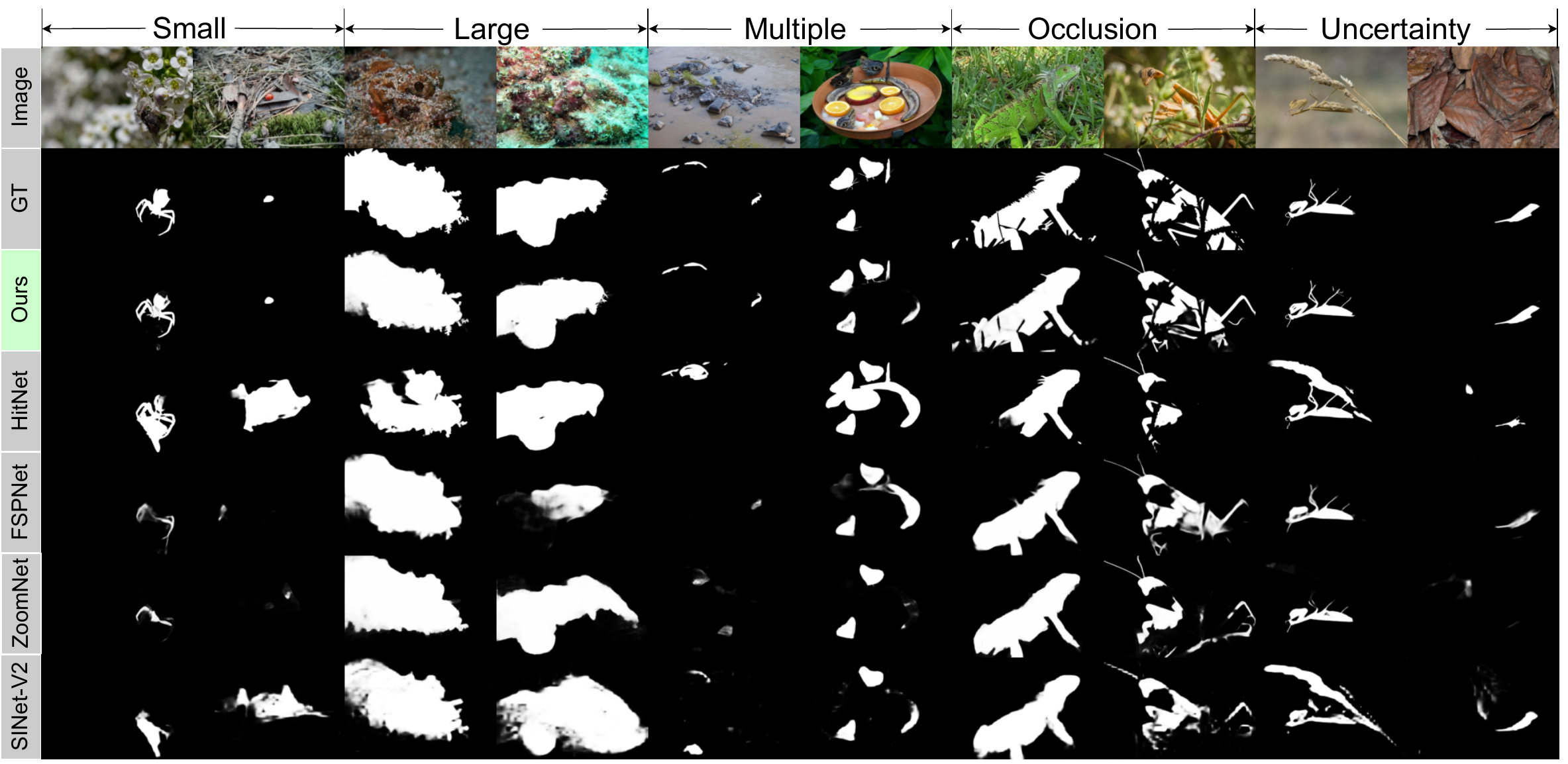}
  \vspace{0.1cm}
  \caption{\textbf{Qualitative comparison with baseline models on various types of camouflaged objects.} Our model effectively captures diverse ranges of camouflages in the datasets.}
  \label{fig:qualitative_results}
\end{figure*}

\vspace{0.1cm} \noindent
\textbf{Qualitative Comparison.}
\cref{fig:qualitative_results} illustrates a few examples on COD10K.
Our model successfully captures both the overall structure and the fine details, such as slim legs or complex edges, compared to other models.
Additionally, \proposal{} captures small objects accurately, and at the same time, accurately covers the entirety of the object for big objects.
In spite of occlusion, \proposal{} accurately captures the target object, while other models often include such occluded parts.
Even for objects with complex shapes, our model is able to capture fine details, while other models tend to inaccurately capture such details.

\subsection{Ablation Study \& Visualization}
\label{subsec:ablation}

\begin{table}
\centering
\footnotesize
\renewcommand{\tabcolsep}{4pt}
\begin{tabular}{l|cccc}
    \toprule
    {\textbf{Decoders}} &
    ${S_\alpha}$ ↑ & ${F_\beta^w}$ ↑ & ${E_\phi}$ ↑ & ${M}$ ↓ 
    \\
    \midrule
    Base              
    & 0.890 & 0.815 & 0.938 & 0.021 
    \\
    Base + Enrich               
    & 0.899 & 0.836 & 0.946 & 0.019
    \\
    Base + Retouch    
    & 0.896 & 0.825 & 0.935 & 0.020
    \\
    \textbf{\proposal\ (ours)}   
    & {\bf 0.904} & {\bf 0.845} & {\bf 0.948} & {\bf 0.018}
    \\
    \bottomrule
\end{tabular}
\vspace{0.2cm}
\caption{\textbf{Ablation study on decoding structure of ENTO on COD10K.} }
\label{tab:decoder_ablation}
\end{table}

\begin{figure}
    \centering
    \begin{minipage}{0.55\textwidth}
        \centering
        \includegraphics[width=\linewidth]{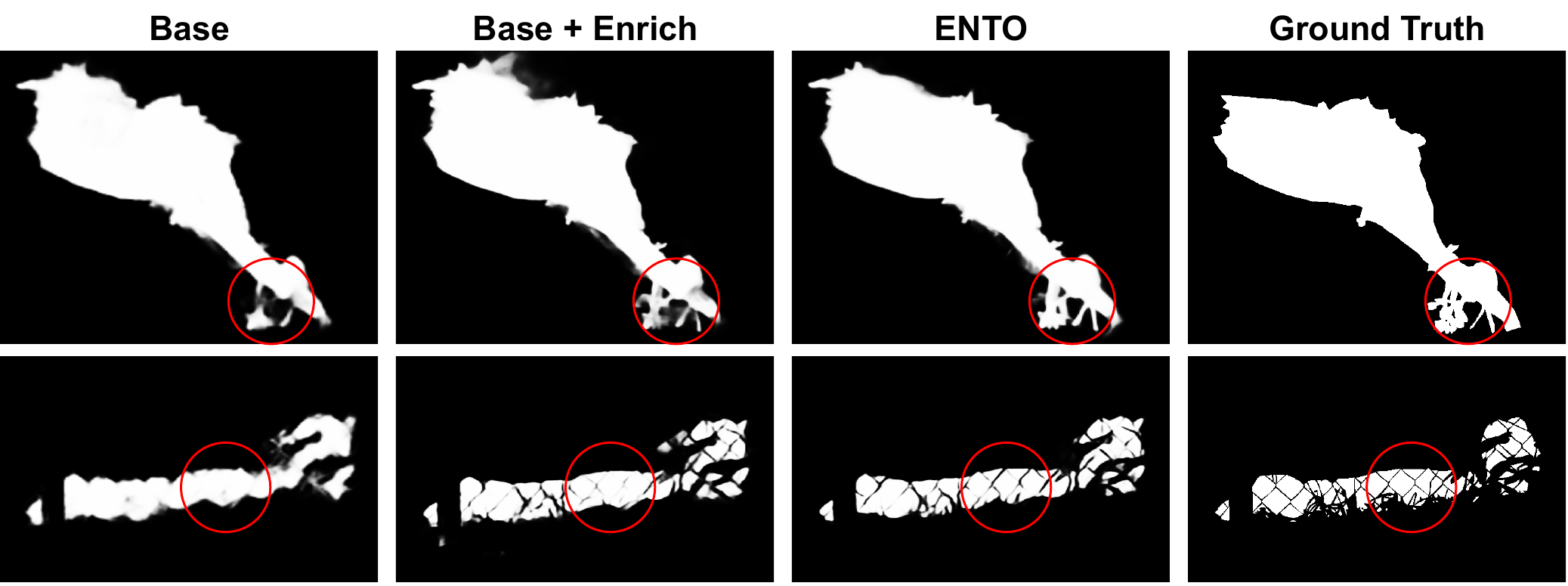}
        \vspace{0.1cm}
        \caption{\textbf{Segmentation maps with ENTO components.} Enrich Decoder adds missing regions using COD-specific features, and Retouch Decoder finetunes the details near the edges.}
        \label{fig:map_examples}
    \end{minipage}\hfill
    \begin{minipage}{0.43\textwidth}
        \includegraphics[width=\linewidth]{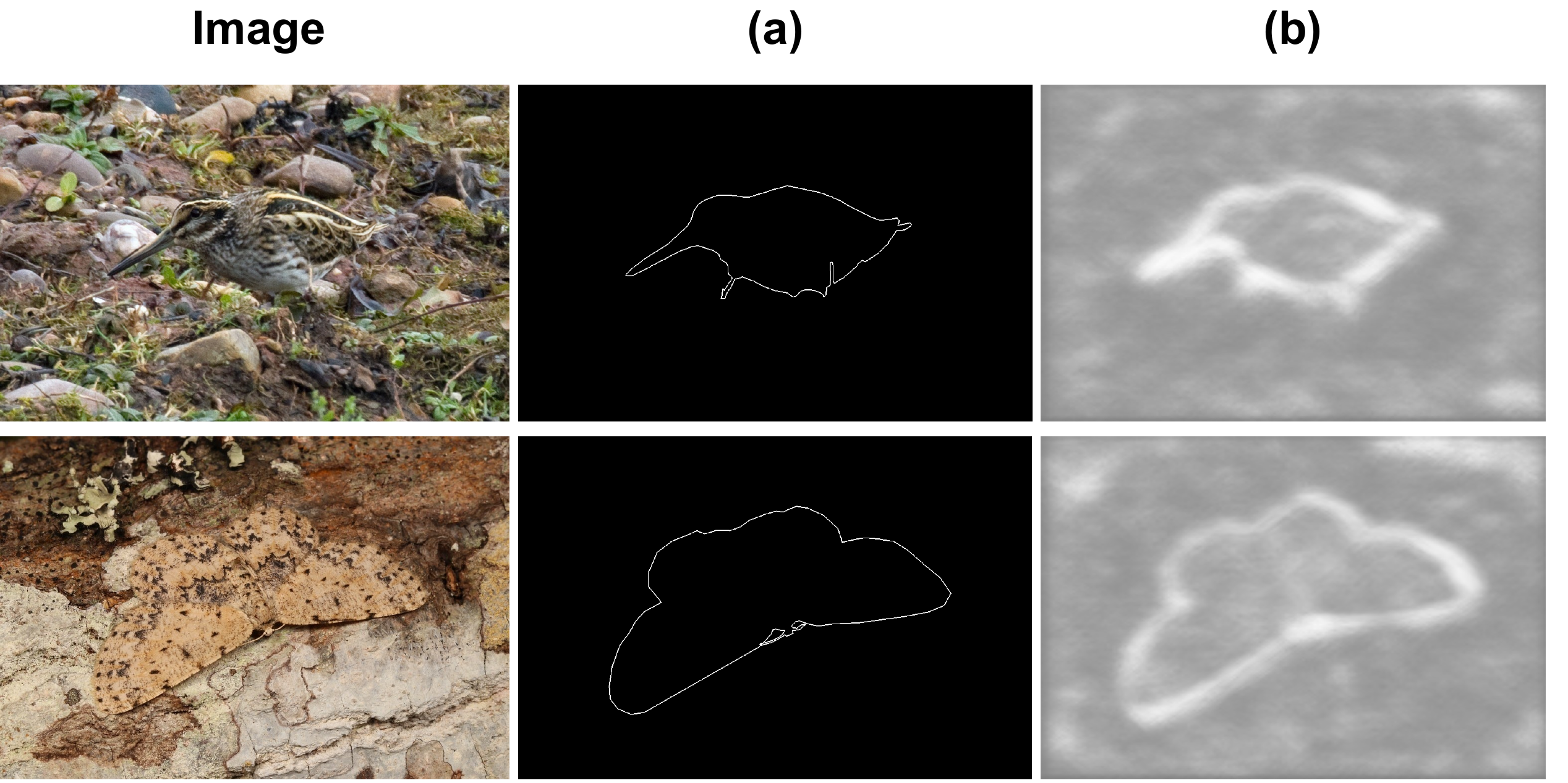}
        \vspace{0.1cm}
        \caption{\textbf{Illustration of Spatial Attention by \postdec.} (a) ground truth of the edge, (b) spatial attention by our SAB.}
        \label{fig:SAB_examples}
    \end{minipage}
\end{figure}

We evaluate the effects of the proposed components by measuring the performance improvement over the \maindec\ alone.
\cref{tab:decoder_ablation} verifies on COD10K that adding each component improves the performance.
As expected, the best performance is achieved with both components, showing that they are complementary.
In \cref{fig:map_examples}, we illustrate that adding the \predec\ recovers the details missing in segmentation map produced by base decoder and adding the \postdec\ sharpens these details to correctly match the true map.

\cref{fig:SAB_examples} illustrates the visual representation of spatial attention in our \postdec.
Notably, the attention scores mainly concentrate on the object boundaries, showing its predominant focus in refining these regions.
In the final phase of decoding, \postdec\ effectively enhances intricate details regarding these edges and boundaries.

\section{Summary}
\label{sec:summary}

We present a novel decoding architecture \proposal\ to utilize high-resolution information extraction and address the complex boundary structures for camouflaged object detection.
We propose \predec\ and \postdec\ applicable to the base COD decoder, emphasizing feature channels beneficial for high-resolution segmentation by channel-wise attention and refining the segmentation map from the base decoder by spatial attention focusing on the edges and fine details, respectively.
Our decoding architecture can be combined with different encoder backbones and shows superior results with Transformer backbones.

\clearpage

\section*{Acknowledgements}

\noindent
This work was supported by the New Faculty Startup Fund from Seoul National University, by Samsung Electronics Co., Ltd (IO230414-05943-01, RAJ0123ZZ-80SD), by Youlchon Foundation (Nongshim Corp.), and by National Research Foundation (NRF) grants (No.
2021H1D3A2A03038607/50\%, RS-2024-00336576/10\%, RS-
2023-00222663/5\%) and Institute for Information \& communication Technology Planning \& evaluation (IITP) grants (No. RS-2024-00353131/25\%, RS-2022-II220264/10\%), funded by the government of Korea.

\bibliography{main}

\begin{thebibliography}{39}
\providecommand{\natexlab}[1]{#1}
\providecommand{\url}[1]{\texttt{#1}}
\expandafter\ifx\csname urlstyle\endcsname\relax
  \providecommand{\doi}[1]{doi: #1}\else
  \providecommand{\doi}{doi: \begingroup \urlstyle{rm}\Url}\fi

\bibitem[Bhajantri and Nagabhushan(2006)]{bhajantri2006camouflage}
Nagappa~U Bhajantri and P~Nagabhushan.
\newblock Camouflage defect identification: a novel approach.
\newblock In \emph{Proc. of the International Conference on Information Technology (ICIT)}, 2006.

\bibitem[Chen et~al.(2022)Chen, Xiao, Hu, Zhang, and Wang]{chen2022bgnet}
Tianyou Chen, Jin Xiao, Xiaoguang Hu, Guofeng Zhang, and Shaojie Wang.
\newblock Boundary-guided network for camouflaged object detection.
\newblock \emph{Knowledge-Based Systems}, 248:\penalty0 108901, 2022.

\bibitem[Dosovitskiy et~al.(2020)Dosovitskiy, Beyer, Kolesnikov, Weissenborn, Zhai, Unterthiner, Dehghani, Minderer, Heigold, Gelly, et~al.]{dosovitskiy2020vit}
Alexey Dosovitskiy, Lucas Beyer, Alexander Kolesnikov, Dirk Weissenborn, Xiaohua Zhai, Thomas Unterthiner, Mostafa Dehghani, Matthias Minderer, Georg Heigold, Sylvain Gelly, et~al.
\newblock An image is worth 16x16 words: Transformers for image recognition at scale.
\newblock In \emph{ICLR}, 2020.

\bibitem[Fan et~al.(2017)Fan, Cheng, Liu, Li, and Borji]{fan2017structure}
Deng-Ping Fan, Ming-Ming Cheng, Yun Liu, Tao Li, and Ali Borji.
\newblock Structure-measure: A new way to evaluate foreground maps.
\newblock In \emph{ICCV}, 2017.

\bibitem[Fan et~al.(2020{\natexlab{a}})Fan, Ji, Sun, Cheng, Shen, and Shao]{fan2020sinet}
Deng-Ping Fan, Ge-Peng Ji, Guolei Sun, Ming-Ming Cheng, Jianbing Shen, and Ling Shao.
\newblock Camouflaged object detection.
\newblock In \emph{CVPR}, 2020{\natexlab{a}}.

\bibitem[Fan et~al.(2020{\natexlab{b}})Fan, Ji, Zhou, Chen, Fu, Shen, and Shao]{fan2020pranet}
Deng-Ping Fan, Ge-Peng Ji, Tao Zhou, Geng Chen, Huazhu Fu, Jianbing Shen, and Ling Shao.
\newblock Pra{N}et: Parallel reverse attention network for polyp segmentation.
\newblock In \emph{Proc. of the Medical Image Computing and Computer Assisted Intervention (MICCAI)}, 2020{\natexlab{b}}.

\bibitem[Fan et~al.(2021{\natexlab{a}})Fan, Ji, Cheng, and Shao]{fan2021sinetv2}
Deng-Ping Fan, Ge-Peng Ji, Ming-Ming Cheng, and Ling Shao.
\newblock Concealed object detection.
\newblock \emph{IEEE Transactions on Pattern Analysis and Machine Intelligence}, 44\penalty0 (10):\penalty0 6024--6042, 2021{\natexlab{a}}.

\bibitem[Fan et~al.(2021{\natexlab{b}})Fan, Ji, Qin, and Cheng]{fan2021emeasure}
Deng-Ping Fan, Ge-Peng Ji, Xuebin Qin, and Ming-Ming Cheng.
\newblock Cognitive vision inspired object segmentation metric and loss function.
\newblock \emph{Scientia Sinica Informationis}, 6\penalty0 (6), 2021{\natexlab{b}}.

\bibitem[Ha et~al.(2024)Ha, Kim, Kim, Lee, Lee, and Lee]{ha2024nemo}
Seongsu Ha, Chaeyun Kim, Donghwa Kim, Junho Lee, Sangho Lee, and Joonseok Lee.
\newblock Finding {NeMo}: Negative-mined mosaic augmentation for referring image segmentation.
\newblock In \emph{ECCV}, 2024.

\bibitem[He et~al.(2023)He, Li, Zhang, Tang, Zhang, Guo, and Li]{he2023fedder}
Chunming He, Kai Li, Yachao Zhang, Longxiang Tang, Yulun Zhang, Zhenhua Guo, and Xiu Li.
\newblock Camouflaged object detection with feature decomposition and edge reconstruction.
\newblock In \emph{CVPR}, 2023.

\bibitem[He et~al.(2016)He, Zhang, Ren, and Sun]{he2016resnet}
Kaiming He, Xiangyu Zhang, Shaoqing Ren, and Jian Sun.
\newblock Deep residual learning for image recognition.
\newblock In \emph{CVPR}, 2016.

\bibitem[Hu et~al.(2023)Hu, Wang, Qin, Dai, Ren, Luo, Tai, and Shao]{hu2023hitnet}
Xiaobin Hu, Shuo Wang, Xuebin Qin, Hang Dai, Wenqi Ren, Donghao Luo, Ying Tai, and Ling Shao.
\newblock High-resolution iterative feedback network for camouflaged object detection.
\newblock In \emph{AAAI}, 2023.

\bibitem[Huang et~al.(2023)Huang, Dai, Xiang, Wang, Chen, Qin, and Xiong]{huang2023fspnet}
Zhou Huang, Hang Dai, Tian-Zhu Xiang, Shuo Wang, Huai-Xin Chen, Jie Qin, and Huan Xiong.
\newblock Feature shrinkage pyramid for camouflaged object detection with transformers.
\newblock In \emph{CVPR}, 2023.

\bibitem[Jia et~al.(2022)Jia, Yao, Liu, Fan, Liu, and Luo]{jia2022segmar}
Qi~Jia, Shuilian Yao, Yu~Liu, Xin Fan, Risheng Liu, and Zhongxuan Luo.
\newblock Segment, magnify and reiterate: Detecting camouflaged objects the hard way.
\newblock In \emph{CVPR}, 2022.

\bibitem[Kim et~al.(2019)Kim, Kim, Lee, Yoon, Kahou, Kashinath, and Prabhat]{kim2019deep}
Sookyung Kim, Hyojin Kim, Joonseok Lee, Sangwoong Yoon, Samira~Ebrahimi Kahou, Karthik Kashinath, and Mr~Prabhat.
\newblock Deep-hurricane-tracker: Tracking and forecasting extreme climate events.
\newblock In \emph{WACV}, 2019.

\bibitem[Le et~al.(2019)Le, Nguyen, Nie, Tran, and Sugimoto]{le2019camo}
Trung-Nghia Le, Tam~V Nguyen, Zhongliang Nie, Minh-Triet Tran, and Akihiro Sugimoto.
\newblock Anabranch network for camouflaged object segmentation.
\newblock \emph{Computer vision and image understanding}, 184:\penalty0 45--56, 2019.

\bibitem[Li et~al.(2021)Li, Zhang, Lv, Liu, Zhang, and Dai]{li2021ujsc}
Aixuan Li, Jing Zhang, Yunqiu Lv, Bowen Liu, Tong Zhang, and Yuchao Dai.
\newblock Uncertainty-aware joint salient object and camouflaged object detection.
\newblock In \emph{CVPR}, 2021.

\bibitem[Liu and Di(2023)]{liu2023mhnet}
Maozhen Liu and Xiaoguang Di.
\newblock Extraordinary {MHNet}: Military high-level camouflage object detection network and dataset.
\newblock \emph{Neurocomputing}, page 126466, 2023.

\bibitem[Liu et~al.(2022)Liu, Zhang, Tan, and Wu]{liu2022dtinet}
Zhengyi Liu, Zhili Zhang, Yacheng Tan, and Wei Wu.
\newblock Boosting camouflaged object detection with dual-task interactive transformer.
\newblock In \emph{Proc. of the International Conference on Pattern Recognition (ICPR)}, 2022.

\bibitem[Lv et~al.(2021)Lv, Zhang, Dai, Li, Liu, Barnes, and Fan]{lv2021slsr}
Yunqiu Lv, Jing Zhang, Yuchao Dai, Aixuan Li, Bowen Liu, Nick Barnes, and Deng-Ping Fan.
\newblock Simultaneously localize, segment and rank the camouflaged objects.
\newblock In \emph{CVPR}, 2021.

\bibitem[Lyu et~al.(2021)Lyu, Zhang, Dai, Li, Liu, Barnes, and Fan]{lyu2021rank}
Yunqiu Lyu, Jing Zhang, Yuchao Dai, Aixuan Li, Bowen Liu, Nick Barnes, and Deng-Ping Fan.
\newblock Simultaneously localize, segment and rank the camouflaged objects.
\newblock In \emph{CVPR}, 2021.

\bibitem[Margolin et~al.(2014)Margolin, Zelnik-Manor, and Tal]{margolin2014wfmeasure}
Ran Margolin, Lihi Zelnik-Manor, and Ayellet Tal.
\newblock How to evaluate foreground maps?
\newblock In \emph{CVPR}, 2014.

\bibitem[Mei et~al.(2021)Mei, Ji, Wei, Yang, Wei, and Fan]{mei2021pfnet}
Haiyang Mei, Ge-Peng Ji, Ziqi Wei, Xin Yang, Xiaopeng Wei, and Deng-Ping Fan.
\newblock Camouflaged object segmentation with distraction mining.
\newblock In \emph{CVPR}, 2021.

\bibitem[Merilaita et~al.(2017)Merilaita, Scott-Samuel, and Cuthill]{merilaita2017howcamouflage}
Sami Merilaita, Nicholas~E Scott-Samuel, and Innes~C Cuthill.
\newblock How camouflage works.
\newblock \emph{Philosophical Transactions of the Royal Society B: Biological Sciences}, 372\penalty0 (1724):\penalty0 20160341, 2017.

\bibitem[Pang et~al.(2022)Pang, Zhao, Xiang, Zhang, and Lu]{pang2022zoomnet}
Youwei Pang, Xiaoqi Zhao, Tian-Zhu Xiang, Lihe Zhang, and Huchuan Lu.
\newblock Zoom in and out: A mixed-scale triplet network for camouflaged object detection.
\newblock In \emph{CVPR}, 2022.

\bibitem[Parra-Mora et~al.(2022)Parra-Mora, Caza{\~n}as-Gord{\'o}n, and da~Silva~Cruz]{parra2022inputeffect}
Esther Parra-Mora, Alex Caza{\~n}as-Gord{\'o}n, and Lu{\'\i}s~A da~Silva~Cruz.
\newblock The effect of input size in deep learning semantic segmentation.
\newblock In \emph{Proc. of the IEEE Ecuador Technical Chapters Meeting (ETCM)}, 2022.

\bibitem[Perazzi et~al.(2012)Perazzi, Kr{\"a}henb{\"u}hl, Pritch, and Hornung]{perazzi2012mae}
Federico Perazzi, Philipp Kr{\"a}henb{\"u}hl, Yael Pritch, and Alexander Hornung.
\newblock Saliency filters: Contrast based filtering for salient region detection.
\newblock In \emph{CVPR}, 2012.

\bibitem[Qin et~al.(2019)Qin, Zhang, Huang, Gao, Dehghan, and Jagersand]{qin2019basnet}
Xuebin Qin, Zichen Zhang, Chenyang Huang, Chao Gao, Masood Dehghan, and Martin Jagersand.
\newblock Bas{N}et: Boundary-aware salient object detection.
\newblock In \emph{CVPR}, 2019.

\bibitem[Ronneberger et~al.(2015)Ronneberger, Fischer, and Brox]{ronneberger2015unet}
Olaf Ronneberger, Philipp Fischer, and Thomas Brox.
\newblock U-{N}et: Convolutional networks for biomedical image segmentation.
\newblock In \emph{Proc. of the Medical Image Computing and Computer-Assisted Intervention (MICCAI)}, 2015.

\bibitem[Sun et~al.(2021)Sun, Chen, Zhou, Zhang, and Liu]{sun2021c2fnet}
Yujia Sun, Geng Chen, Tao Zhou, Yi~Zhang, and Nian Liu.
\newblock Context-aware cross-level fusion network for camouflaged object detection.
\newblock \emph{arXiv:2105.12555}, 2021.

\bibitem[Vaswani et~al.(2017)Vaswani, Shazeer, Parmar, Uszkoreit, Jones, Gomez, Kaiser, and Polosukhin]{vaswani2017attention}
Ashish Vaswani, Noam Shazeer, Niki Parmar, Jakob Uszkoreit, Llion Jones, Aidan~N Gomez, {\L}ukasz Kaiser, and Illia Polosukhin.
\newblock Attention is all you need.
\newblock \emph{NeurIPS}, 30, 2017.

\bibitem[Wang et~al.(2022)Wang, Xie, Li, Fan, Song, Liang, Lu, Luo, and Shao]{wang2022pvt}
Wenhai Wang, Enze Xie, Xiang Li, Deng-Ping Fan, Kaitao Song, Ding Liang, Tong Lu, Ping Luo, and Ling Shao.
\newblock {PVT} v2: Improved baselines with pyramid vision transformer.
\newblock \emph{Computational Visual Media}, 8\penalty0 (3):\penalty0 415--424, 2022.

\bibitem[Wei et~al.(2020)Wei, Wang, and Huang]{wei2020f3net}
Jun Wei, Shuhui Wang, and Qingming Huang.
\newblock F$^3${N}et: fusion, feedback and focus for salient object detection.
\newblock In \emph{AAAI}, 2020.

\bibitem[Xing et~al.(2023)Xing, Wang, Wei, Tang, Gao, and Zhang]{xing2023sarnet}
Haozhe Xing, Yan Wang, Xujun Wei, Hao Tang, Shuyong Gao, and Wenqiang Zhang.
\newblock Go closer to see better: Camouflaged object detection via object area amplification and figure-ground conversion.
\newblock \emph{IEEE Transactions on Circuits and Systems for Video Technology}, 2023.

\bibitem[Zeng et~al.(2019)Zeng, Zhang, Zhang, Lin, and Lu]{zeng2019towards}
Yi~Zeng, Pingping Zhang, Jianming Zhang, Zhe Lin, and Huchuan Lu.
\newblock Towards high-resolution salient object detection.
\newblock In \emph{ICCV}, 2019.

\bibitem[Zhai et~al.(2021)Zhai, Li, Yang, Chen, Cheng, and Fan]{zhai2021mgl}
Qiang Zhai, Xin Li, Fan Yang, Chenglizhao Chen, Hong Cheng, and Deng-Ping Fan.
\newblock Mutual graph learning for camouflaged object detection.
\newblock In \emph{CVPR}, 2021.

\bibitem[Zhang et~al.(2022)Zhang, Xu, Piao, Shi, Lin, and Lu]{zhang2022preynet}
Miao Zhang, Shuang Xu, Yongri Piao, Dongxiang Shi, Shusen Lin, and Huchuan Lu.
\newblock Prey{N}et: Preying on camouflaged objects.
\newblock In \emph{ACM MM}, 2022.

\bibitem[Zhang et~al.(2018)Zhang, Li, Li, Wang, Zhong, and Fu]{zhang2018cab}
Yulun Zhang, Kunpeng Li, Kai Li, Lichen Wang, Bineng Zhong, and Yun Fu.
\newblock Image super-resolution using very deep residual channel attention networks.
\newblock In \emph{ECCV}, 2018.

\bibitem[Zhu et~al.(2022)Zhu, Li, Xie, Yan, Liang, Chen, Wei, and Qin]{zhu2022bsanet}
Hongwei Zhu, Peng Li, Haoran Xie, Xuefeng Yan, Dong Liang, Dapeng Chen, Mingqiang Wei, and Jing Qin.
\newblock I can find you! boundary-guided separated attention network for camouflaged object detection.
\newblock In \emph{AAAI}, 2022.

\end{thebibliography}

\clearpage

\setcounter{page}{1}

\section*{Appendix}

\appendix

\pagenumbering{roman}
\renewcommand\thetable{\Roman{table}}
\renewcommand\thefigure{\Roman{figure}}
\setcounter{table}{0}
\setcounter{figure}{0}


\section{Group Attention Block}
\label{sec:gab}

\begin{figure}[ht]
  \centering
  \includegraphics[width=0.8\linewidth]{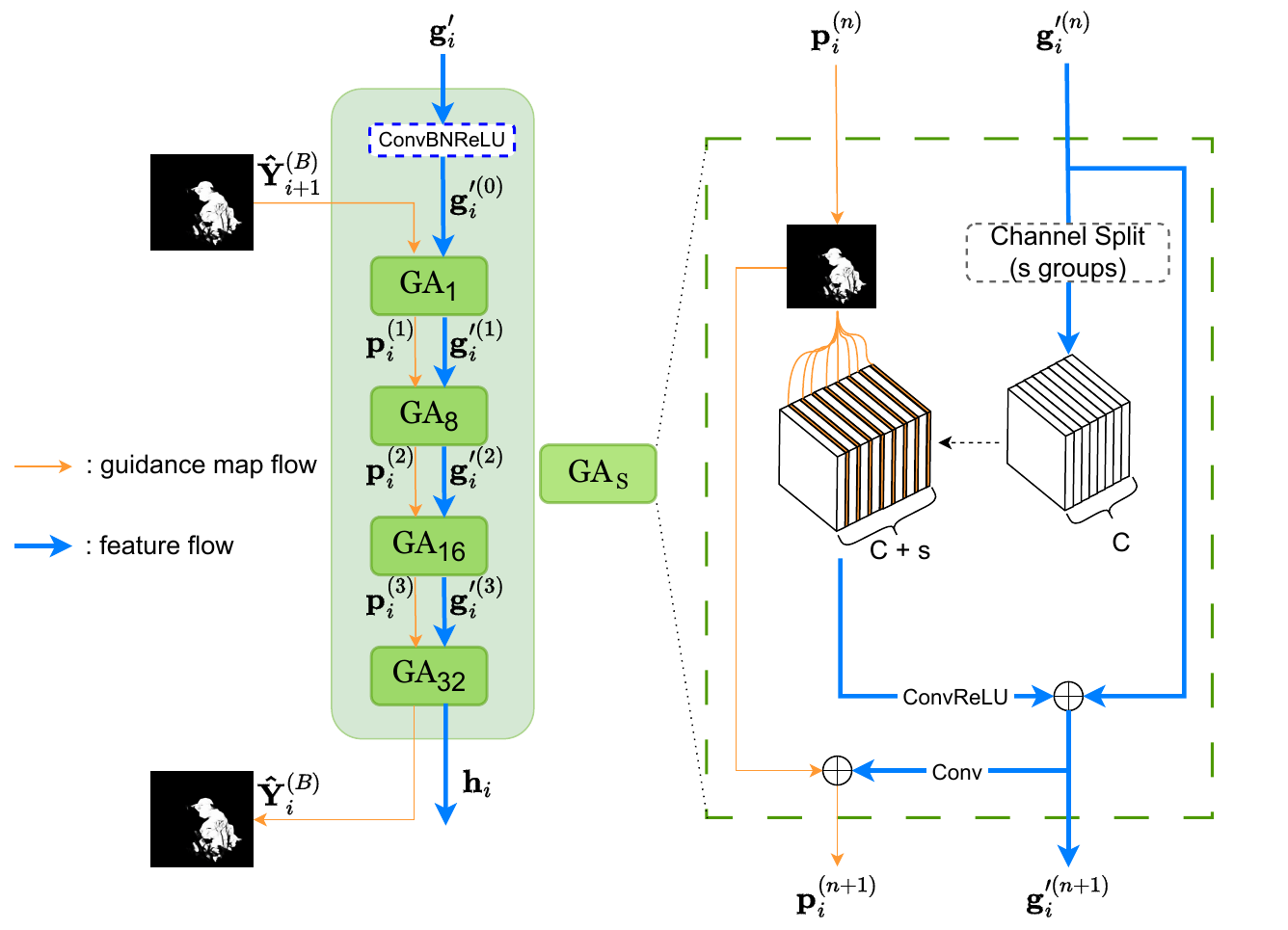}
  \caption{\textbf{Group Attention Block (GAB)}.}
  \label{fig:GAB}
\end{figure}

The Group Attention Block (GAB) is a residual learning process employing the group guidance operation, suggested by Fan et al.~\cite{fan2021sinetv2}.
It focuses on more important information about objects through attention with guidance from the prior segmentation map and gradually improves the map through a sequence of four group attention ($\text{GA}$) operations.  
Each $\text{GA}_s$ operation, as shown in \cref{fig:GAB}, consists of 3 steps: 1) splitting the input feature $\mathbf{g'}_i^{(n)}$ into multiple ($s \in \{1, 8, 16, 32\}$) groups along the channel, 2) concatenating the guidance map $\mathbf{p}_i^{(n)}$ among the split features $\mathbf{g'}_{i,j}^{(n)}$, where $j = 1, ..., s$, and three finally producing an improved guidance map and feature map with convolution operations and residual connection.

Formally, the $n$\textsuperscript{th} GA operation of the GAB at level $i$ is given by
\begin{align}
  \text{1) } & \mathbf{g'}_{i}^{(n)} \rightarrow \begin{Bmatrix} \mathbf{g'}_{i,1}^{(n)}, \mathbf{g'}_{i,2}^{(n)}, ..., \mathbf{g'}_{i,s}^{(n)} \end{Bmatrix} \\
  \text{2) } & 
  \mathbf{g'p}_{i}^{(n)} = \mathtt{Cat}\begin{bmatrix} \mathbf{g'}_{i,1}^{(n)}, \mathbf{p}_{i}^{(n)}, ... , \mathbf{g'}_{i,s}^{(n)}, \mathbf{p}_{i}^{(n)} \end{bmatrix}  \\
  \text{3) } & 
   \mathbf{g'}_{i}^{(n+1)} = \mathbf{g'}_{i}^{(n)} \oplus \mathtt{ReLU}\circ\mathtt{Conv3}(\mathbf{g'p}_{i}^{(n)}),\\
   & \mathbf{p}_{i}^{(n+1)} = \mathbf{p}_{i}^{(n)} \oplus \mathtt{Conv3}(\mathbf{g'}_{i}^{(n+1)} )
  \label{GA}
\end{align}

\noindent where  $i = 1, ..., L$, $n \in \{0,1,2,3\}$,  $\mathbf{g'}_{i}^{(0)} = \mathbf{g'}_{i}$, $\mathbf{g'}_{i}^{(4)} = \mathbf{h}_{i}$, $\mathbf{p}^{(0)}_i=\mathbf{\hat{Y}}^{(B)}_{i+1}$, and $\mathbf{p}_i^{(4)}=\mathbf{\hat{Y}}^{(B)}_{i}$. $\mathtt{Cat}$ indicates concatenation along the channel dimension.
The convolution operations reduce the channel number from $C+s$ to $C$ for the feature maps and from $C+s$ to 1 for guidance maps, which in turn becomes the guidance for the next GA operation. 
For the first GAB stage, a coarse map from the \predec\ ($\mathbf{\hat{Y}}^{(E)}$) is used as the guidance segmentation map.
Each GAB iteratively refines the output segmentation map, which is used as guidance for the first GA operation at the next GAB stage.

\section{Parameter Size Comparison}
\label{sec:computational_comparison}

Comparing the parameter size of architectures is essential to justify that the performance increase is not merely due to larger model capacity. To ascertain that the performance of ENTO is not merely due to an increase in parameter size, we conduct a comparative analysis with parameter sizes of other models, presented in \cref{tab:decoder_parameter}. As ENTO is versatile and can adapt to different backbone encoders, we choose to only compare the decoder parameter sizes, by leaving out the backbone encoder size for all models. The results reveal that our decoder parameter, at 4.17M, is smaller than all other models except SINet-V2. This indicates that the enhancement in performance of our method is not a consequence of increased parameters but rather the result of an efficient architectural design.

\begin{table}[ht]
\centering 
\begin{tabular}{@{}l|c|cccc@{}}
\toprule
\textbf{Model} & \textbf{Decoder Params} & ${S_\alpha} \uparrow$ & ${F_\beta^w} \uparrow$ & ${E_\phi} \uparrow$ & ${M} \downarrow$ \\
\midrule 
SINet          & 23.35M                   & 0.808 & 0.723 & 0.871 & 0.058 \\
MGL-R          & 42.04M                   & 0.833 & 0.740  & 0.867 & 0.052 \\
PFNet          & 20.90M                    & 0.829 & 0.745 & 0.888 & 0.053 \\
SINet-V2       & 1.69M                    & 0.847 & 0.770  & 0.903 & 0.048 \\
SegMaR         & 42.44M                   & 0.841 & 0.781 & 0.896 & 0.046 \\
ZoomNet        & 6.78M                    & 0.853 & 0.784 & 0.912 & 0.043 \\
FSPNet        & 188.15M                    & 0.879 & 0.816 & 0.915 & 0.048 \\
FEDER-R2N        & 18.52M                    & 0.862 & - & 0.913 & 0.042 \\
\textbf{ENTO(Ours)}     & 4.17M                    & \textbf{0.904} & \textbf{0.864} & \textbf{0.942} & \textbf{0.029} \\
\bottomrule
\end{tabular}
\vspace{0.2cm}
\caption{\textbf{Comparative analysis of decoder parameter sizes across different models.} Performance metrics are evaluated on the NC4K dataset.}
\label{tab:decoder_parameter}
\end{table}

\section{Implementation Details}
\label{sec:impl_details}

All input images are resized to the desired input size and augmented by random flipping and rotation.  
\proposal\ can adopt various backbones for the encoder.
We report the main results in \cref{tab:all_results} using PVTv2-B4~\cite{wang2022pvt} pretrained on ImageNet-1K and with $768 \times 768$ input size. 
For fair comparison with previous state-of-the-art models, and to demonstrate the versatility of our architecture with any encoder, we also report ablation results using a comparable backbone and input resolution in \cref{tab:impartial_comparison}.
For training our model, we set the learning rate to 0.01 by default, except for the backbone (0.001).
We linearly warm up the learning rate for the first half of training and decrease it to 0 for the rest.
We use SGD optimizer with 0.9 momentum and 0.0005 weight decay.
We train our model up to 100 epochs and set the batch size to 16.
We conduct experiments on a single NVIDIA A100 GPU, taking about 10 hours to train a model.
The Implementation details for other backbones are provided in \cref{sec:encoder_setting}.

\section{Impartial Comparison Setting}
\label{sec:encoder_setting}


\begin{table*}[ht]
\centering
\footnotesize
\renewcommand{\tabcolsep}{4pt}
\resizebox{\textwidth}{!}{%
\begin{tabular}{lcccc|c|c|c|c|cc|c|c}
\toprule

\multicolumn{1}{c|}{\multirow{2}{*}{\textbf{Encoder Backbone}}} & \multicolumn{1}{c|}{\multirow{2}{*}{\textbf{Resolution}}}  & \multicolumn{1}{c|}{\multirow{2}{*}{\textbf{Best Baseline}}} & \multicolumn{1}{c|}{\textbf{Feature}} & \multicolumn{1}{c|}{\multirow{2}{*}{\textbf{Channels}}} & \multicolumn{1}{c|}{\multirow{1}{*}{\textbf{CABs}}} & \multicolumn{1}{c|}{\multirow{1}{*}{\textbf{SABs}}} & \multicolumn{2}{c|}{\textbf{Batch Size}} & \multicolumn{1}{c|}{\textbf{Learning}}  & \multicolumn{1}{c}{{\textbf{Training}}}
\\ 
\multicolumn{1}{c|}{} & \multicolumn{1}{c|}{} & \multicolumn{1}{c|}{} & \multicolumn{1}{c|}{\textbf{Level ($L$)}} & \multicolumn{1}{c|}{} & \multicolumn{1}{c|}{\textbf{per Level}} & \multicolumn{1}{c|}{\textbf{per Level}} & Ours & Base &\multicolumn{1}{c|}{\textbf{Rate}} &  \multicolumn{1}{c}{\textbf{Epohcs}}  \\ \midrule

\multicolumn{1}{l|}{\multirow{2}{*}{PVTv2-B2~\cite{wang2022pvt}}} & \multicolumn{1}{c|}{$352\times352$}& \multicolumn{1}{c|}{HitNet~\cite{hu2023hitnet}} & \multicolumn{1}{c|}{4} & \multicolumn{1}{c|}{64} & 6 & 6 & 16 & - & \multicolumn{1}{c|}{1e-2} & \multicolumn{1}{c}{100} \\
\multicolumn{1}{c|}{ } & \multicolumn{1}{c|}{$704\times704$} & \multicolumn{1}{c|}{HitNet~\cite{hu2023hitnet}} & \multicolumn{1}{c|}{4} & \multicolumn{1}{c|}{64} & 6 & 6 & 16 & 16 & \multicolumn{1}{c|}{1e-2} & \multicolumn{1}{c}{100} \\ \midrule

\multicolumn{1}{l|}{ViT~\cite{dosovitskiy2020vit}} & \multicolumn{1}{c|}{$384\times384$} & \multicolumn{1}{c|}{FSPNet~\cite{huang2023fspnet}} &
 \multicolumn{1}{c|}{4} & \multicolumn{1}{c|}{64} & 6 & 6 & 16 & 2 & \multicolumn{1}{c|}{1e-2} & \multicolumn{1}{c}{100} \\ \midrule

\multicolumn{1}{l|}{\multirow{2}{*}{Res2Net50~\cite{fan2020sinet}}} & \multicolumn{1}{c|}{$352\times352$}& \multicolumn{1}{c|}{SINet-V2~\cite{fan2021sinetv2}} & \multicolumn{1}{c|}{3} & \multicolumn{1}{c|}{64} & 5 & 5 & 32 & 36 & \multicolumn{1}{c|}{1e-2} & \multicolumn{1}{c}{100} \\

\multicolumn{1}{c|}{ } & \multicolumn{1}{c|}{$384\times384$} & \multicolumn{1}{c|}{FEDER-R2N~\cite{he2023fedder}}  & \multicolumn{1}{c|}{4} & \multicolumn{1}{c|}{96} & 4 & 4 & 32 & 36 & \multicolumn{1}{c|}{1e-2} & \multicolumn{1}{c}{120} \\ \midrule 

\multicolumn{1}{l|}{ResNet50~\cite{he2016resnet}} & \multicolumn{1}{c|}{$576\times576$} & \multicolumn{1}{c|}{ZoomNet~\cite{pang2022zoomnet}} & \multicolumn{1}{c|}{3} & \multicolumn{1}{c|}{64} & 5 & 5 & 8 & 8 & \multicolumn{1}{c|}{1e-2} & \multicolumn{1}{c}{100} \\ \bottomrule

\end{tabular}
}
\vspace{0.2cm}
\caption{\textbf{Experiment settings for impartial comparison.} In all cases, we try to match the level of features and other parameters used in the best baseline setting.}
\label{tab:encoder_settings}
\end{table*}

In \cref{tab:decoder_ablation}, the main text demonstrates the experimental results of our model under various combinations of encoder backbones and resolutions compared to the best-performing baseline models.
Since different backbones extract features in different numbers of layers and channels, we adapt some model architectures or training settings according to them, mostly following the settings of baseline models.
For each backbone, features from the last $L$ layers are used.
\cref{tab:impartial_comparison} shows the optimal settings of our model on each backbone condition according to our experiments.
The channels column indicates the number of feature channels that are matched in the encoder, and the following decoder architecture is also adapted to the input channel.
For backbone training, the learning rate is applied at 1/10 of the specified rate in the table.
For a fair comparison with ZoomNet~\cite{pang2022zoomnet}, we use the highest resolution ($\times1.5$) images among the multi-scale setting on inputs of the ZoomNet.



\section{Impact of Input Resolution}
\label{sec:input_resolution}

We experiment with various resolutions and report the performance of our proposed model in \cref{tab:resolution_ablation}.
We select various resolutions ranging from $352\times352$ to $896\times896$, considering the resolution choices in literature and average image size of the datasets we utilized: COD10K ($740\times963$) and NC4K ($529\times709$). 

\begin{table*}
\centering
\footnotesize
\begin{tabular}{c|cccc|cccc}
    \toprule
    \multicolumn{1}{c|}{ } & \multicolumn{4}{c|}{\textbf{COD10K}} & \multicolumn{4}{c}{\textbf{NC4K}} \\
     & \multicolumn{4}{c|}{(Avg. Size: $740\times963$)} & \multicolumn{4}{c}{(Avg. Size: $529\times709$)} \\
    Resolution & ${S_\alpha}$ ↑ & ${F_\beta^w}$ ↑ & ${E_\phi}$ ↑ & ${M}$ ↓ & ${S_\alpha}$ ↑ & ${F_\beta^w}$ ↑ & ${E_\phi}$ ↑ & ${M}$ ↓ \\
    \midrule
                   
    $352\times352$ & 0.871 & 0.779 & 0.928 & 0.023 & 0.891 & 0.842 & 0.936 & 0.031 \\

    $384\times384$ & 0.877 & 0.791 & 0.931 & 0.022 & 0.897 & 0.850 & 0.939 & \textbf{0.029} \\
                   
    $416\times416$ & 0.881 & 0.799 & 0.937 & 0.021 & 0.898 & 0.852 & 0.940 & \textbf{0.029} \\
        
    $480\times480$ & 0.889 & 0.815 & 0.941 & 0.020 & 0.900 & 0.858 & 0.941 & \textbf{0.029} \\
        
    
    $704\times704$ & 0.901  & 0.841 & 0.949 & \textbf{0.018} & 0.902 & 0.861 & 0.940 & \textbf{0.029} \\

    $\mathbf{768\times768}$ & 0.904 & 0.845 & 0.948 & \textbf{0.018} & \textbf{0.904} & \textbf{0.864} & \textbf{0.942} & \textbf{0.029} \\
    
    $896\times896$ & \textbf{0.905} & \textbf{0.847} & \textbf{0.950} & \textbf{0.018} & 0.902 & 0.861 & 0.940 & \textbf{0.029} \\
    
    \bottomrule
\end{tabular}
\vspace{0.2cm}
\caption{\textbf{Ablation study on different input resolution.} Bolded result shows our choice of input resolution.}
\label{tab:resolution_ablation}
\end{table*}

As shown in \cref{tab:resolution_ablation}, higher resolutions of input images lead to improved performance until the resolution reaches the average size of each dataset.
We observed no performance improvement or even slight degradation when the resolution exceeded the average size.
Taking into account these aspects and computational complexity, we select $768\times768$ as the representative resolution for our full model performance reported in \cref{tab:all_results}.

\section{Analysis on Imperfect Ground Truths in CAMO Dataset}

As evidenced by the performance results in \cref{tab:all_results}, most of the COD models perform the worst in the CAMO \cite{le2019camo} dataset. Similarly, \proposal's performance in the CAMO dataset is far lower than for other test datasets. We shed some light on why this may be so.

\begin{figure*}[t]
  \centering
  \includegraphics[width=1\linewidth]{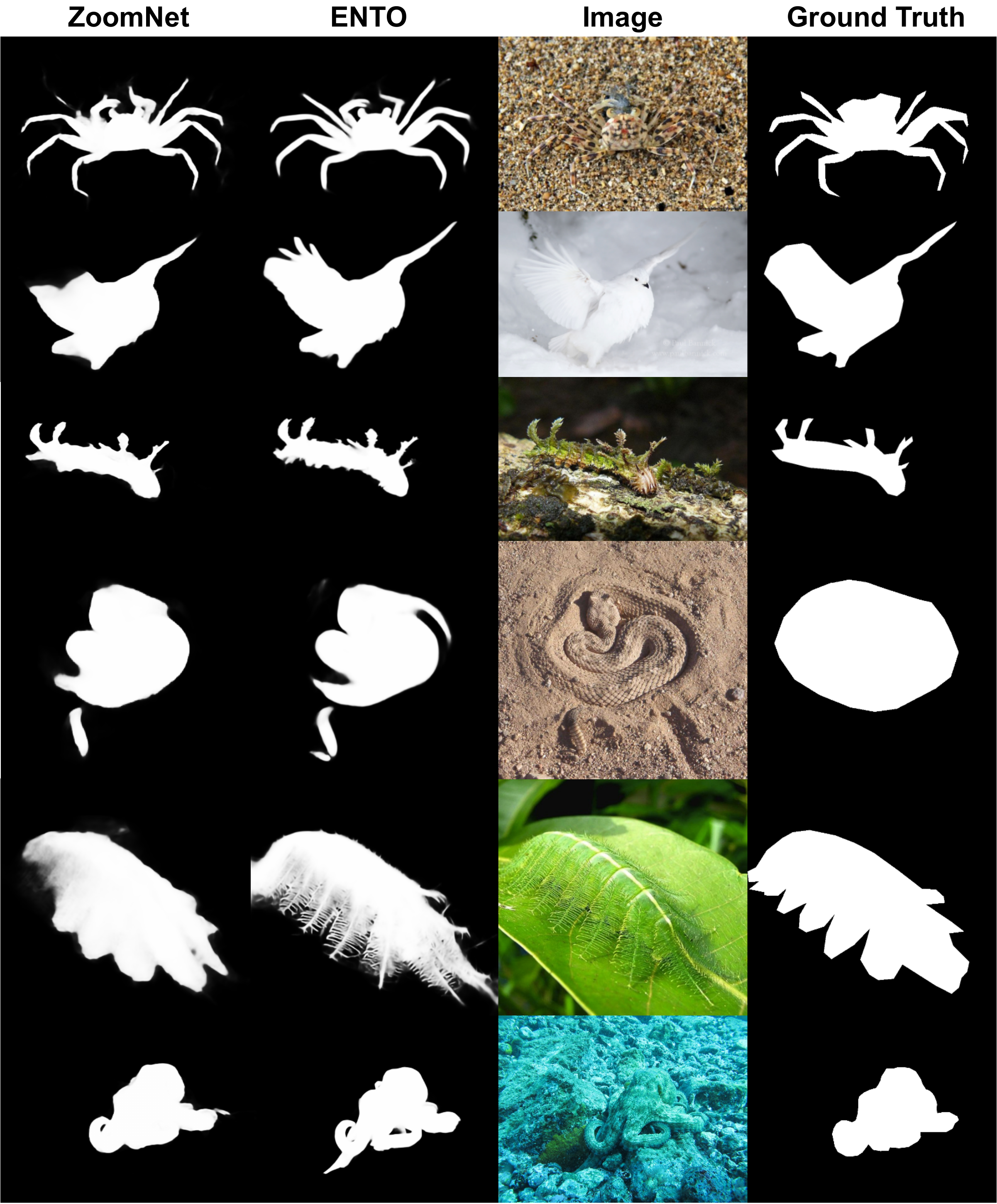}
  \vspace{0.2cm}
  \caption{\textbf{Examples of our model outperforming the ground truths in the CAMO dataset.} The ground truths map neglects details of the object, lump parts of objects together, and miss crucial parts of the objects. Our model is able to capture all these details, while baseline model, ZoomNet~\cite{pang2022zoomnet}, misses such details.}
  \label{fig:camo_ex}
\end{figure*}

As evidenced in \cref{fig:camo_ex}, for some of the images in the CAMO dataset, the ground truth masks are not detailed and lump parts of the object together with the background.
In some cases, crucial parts of an object are not included in the mask as well.
In all such cases, \proposal\ successfully recovers the ``mistakes'' of the ground truths, leading to a more detailed and fine-grained segmentation map. Additionally, compared to baseline model, ZoomNet~\cite{pang2022zoomnet}, ENTO more successfully recovers the missing parts of the objects in the ground truth and generates a more fine-grained segmentation map, matching the actual object.
Such success in the model would have been penalized in the evaluation process, as they do not match the ground truth masks.

\end{document}